\date{}
\documentclass[11pt]{article}

\usepackage[margin=1in]{geometry}
\usepackage{amsmath,amssymb,amsthm}
\usepackage{graphicx}
\usepackage{tikz}
\usepackage{hyperref}
\usepackage{xcolor}
\usepackage{enumitem}
\usepackage{booktabs}
\usepackage{multirow}
\usepackage{cite}

\newtheorem{theorem}{Theorem}[section]
\newtheorem{proposition}[theorem]{Proposition}
\theoremstyle{definition}

\theoremstyle{remark}

\title{\textbf{Computational Challenges in Token Economics:\\Bridging Economic Theory and AI System Design}}

\author{
Ou Wu$^1$, Yingjun Deng$^2$\\
$^1$\textit{Hangzhou Institute for Advanced Study, University of Chinese Academy of Sciences, China}\\
$^2$\textit{Hefei Institutes of Physical Science, Chinese Academy of Sciences, China}\\
\textit{wuou@ucas.ac.cn}, \textit{yingjun.deng@inest.cas.cn}
}

\begin{document}

\maketitle

\begin{abstract}
Token economics has emerged as a useful lens for understanding resource allocation, value creation, and pricing in large language model systems. While recent work has increasingly treated tokens as economic primitives, there remains a substantial gap between high-level economic theory and the computational realities of modern AI infrastructure. This paper identifies and analyzes the key computational challenges that arise when token-economic principles are implemented in real-time inference systems. We argue that computational feasibility is not merely one dimension of token economics, but its governing constraint: these challenges are driven by fundamental tensions among fine-grained valuation, low-latency execution, and allocation optimality under uncertainty. To structure this problem space, we introduce the notion of \textbf{Computational Token Economics} and propose the \textbf{Token Economics Trilemma}---a conditional no-free-lunch principle that captures the inherent trade-offs among granularity, real-time performance, and optimality. We further categorize the main technical challenges into three areas: real-time value accounting, constrained resource allocation, and economic-aware system architecture. Rather than presenting a complete solution, this paper aims to define a research agenda for bridging token economics and AI system design, highlighting open problems at the intersection of computational economics, machine learning systems, and AI infrastructure.

\textbf{Keywords:} Token Economics, Large Language Models, Resource Allocation, Real-time Systems, Computational Economics, AI Infrastructure
\end{abstract}

\section{Introduction}

Large language models (LLMs) have made tokens a central operational unit of modern artificial intelligence systems. Traditionally, tokens were treated mainly as linguistic or computational artifacts. In contemporary LLM services, however, tokens also determine pricing, latency, memory consumption, throughput, and the allocation of accelerator resources. Input tokens consume context-window capacity and key-value (KV) cache; output tokens incur sequential decoding cost; reasoning tokens may be generated internally by the model or hidden behind opaque commercial APIs; and tokens allocated to retrieval, tools, agents, or intermediate reasoning traces can directly affect both service quality and computational cost. This evolution suggests that tokens should be studied not only as model inputs and outputs, but also as economic primitives.

Recent work has begun to formalize this perspective. Chen et al. develop a dual computing--economics perspective on token economics for LLM agents, positioning tokens as production factors, exchange media, and units of account~\cite{chen2026token}. Zhu argues that agentic AI systems should be designed as marginal token allocators, where routers, agents, serving stacks, and training pipelines repeatedly decide where the next unit of tokenized computation should be spent~\cite{zhu2026agentic}. Related studies examine token pricing, optimal token allocation, token futures, and cooperative credit assignment in multi-agent settings~\cite{zhong2025token,smolin2025economics,xing2026token,hua2025shapley}.

These contributions establish the economic vocabulary and structural taxonomy of tokenized AI systems. However, they largely operate under the implicit assumption that token-level valuation, optimization, and coordination can be computed when needed. This paper takes a fundamentally different stance: we argue that \emph{computational feasibility is not merely one dimension of token economics, but its governing constraint}. Once tokens are treated as heterogeneous economic objects whose marginal values must be estimated, allocated, and audited inside latency-critical inference loops, the question is no longer only how tokens \emph{should} be priced or allocated in principle, but how token-economic mechanisms \emph{can} be executed under strict serving constraints. We therefore propose \textbf{Computational Token Economics} as a distinct research area at the intersection of token economics, LLM inference systems, and computational resource allocation.

This emerging literature points to a common theme: tokens are heterogeneous economic objects. Different tokens may have different marginal values, different production costs, and different effects on downstream utility. A short instruction token may determine the behavior of an entire response, whereas a redundant context token may consume memory with limited marginal benefit. A generated token may be costly because of autoregressive decoding, while a hidden reasoning token may influence the quality and cost of a model response without being directly visible to the user. Recent studies on predictable token importance, token-level attribution, utility-aware data pricing, tokenization incentives, and invisible reasoning-token accounting further reinforce the need for explicit token valuation and token accounting~\cite{abdelfattah2025tokenbutler,xiao2025tokenshapley,xu2026utility,artola2025overcharging,sun2025coin}.

The importance of token allocation is further amplified by recent advances in long-context and reasoning-oriented models. Models with very long context windows, such as Gemini 1.5 and Llama 3, expose context length as an increasingly important computational resource~\cite{reid2024gemini15,dubey2024llama3}. Benchmarks and architectural studies show that larger context windows do not automatically imply uniform or effective use of all tokens, motivating more careful treatment of token relevance and context utilization~\cite{hsieh2024ruler,munkhdalai2024infini}. At the same time, reasoning-oriented models and test-time scaling studies suggest that output quality can depend critically on how many tokens are allocated to intermediate reasoning, search, and verification~\cite{deepseek2025r1,team2025kimi,snell2024scaling}. These trends make token allocation not merely a pricing concern, but a key mechanism for controlling quality, latency, and cost.

Despite this progress, there remains a substantial gap between token-economic theory and deployable AI systems. Economic models often assume that valuations can be estimated, objectives can be optimized, and allocation mechanisms can be executed at a timescale suitable for market clearing. Production LLM inference operates under a very different regime. Decisions about batching, routing, prefill--decode scheduling, KV-cache placement, speculative execution, and admission control must be made under strict latency and memory constraints. Moreover, the future value of tokens is uncertain at the time when many system decisions must be made. A mechanism that is attractive from an economic perspective may therefore become infeasible if it requires per-token value estimation, repeated counterfactual evaluation, iterative optimization, or global coordination during online inference.

Recent system studies show why this implementation gap is fundamental rather than incidental. High-throughput LLM serving depends on fine-grained control of batching, prefill--decode scheduling, and KV-cache memory. Sarathi-Serve improves serving efficiency through chunked prefills and decode piggybacking, while DistServe disaggregates prefill and decoding to improve goodput under heterogeneous workloads~\cite{agrawal2024sarathi,zhong2024distserve}. KV-cache compression and selection methods such as KIVI and SnapKV further indicate that not all context tokens should be treated equally from a memory-allocation perspective~\cite{liu2024kivi,li2024snapkv}. Efficient mixture-of-experts and inference-oriented model designs also highlight the increasing pressure to jointly optimize quality and computational cost~\cite{deepseek2024v2}. These results imply that token-level economic mechanisms cannot be designed independently of serving architecture, memory hierarchy, and latency budgets.

The difficulty is especially visible in token valuation. Exact or high-fidelity valuation is appealing because it can support pricing, compression, caching, context selection, agent credit assignment, and quality-aware allocation. However, principled attribution mechanisms are often computationally expensive. Shapley-style methods provide a rigorous notion of marginal contribution, and recent work has applied such ideas to token-level context attribution and cooperative behavior among LLM agents~\cite{xiao2025tokenshapley,hua2025shapley}. Nevertheless, their direct use in online inference requires approximation, amortization, or architectural support. Similarly, predictable token-importance models suggest that token value may be estimated before or during inference~\cite{abdelfattah2025tokenbutler}, but such estimates must be integrated with real-time serving constraints.

We refer to this emerging problem area as \emph{Computational Token Economics}. Its goal is not only to ask how tokens should be valued or priced in principle, but also to study how token-economic mechanisms can be implemented under the computational constraints of real AI infrastructure. This perspective treats latency, memory bandwidth, KV-cache capacity, batching behavior, accelerator utilization, hidden reasoning tokens, verifiability, and uncertainty as first-class components of the economic problem. It therefore complements recent work on token economics by focusing on the computational feasibility of token-level valuation and allocation. Recent work on advantage-aware speculative reasoning and trustless verifiable inference further suggests that token-level economic decisions may interact with reasoning efficiency, verification cost, and service trustworthiness~\cite{maheswaran2025arbitrage,ong2025toploc}.

The central thesis of this paper is that computational token economics is governed by a fundamental three-way tension among granularity, real-time performance, and optimality. Fine-grained token valuation can improve allocation quality, but it increases sensing and optimization overhead. Real-time inference requires local, approximate, and hardware-aware decisions, but these decisions may ignore economically relevant information. Strong allocation optimality typically requires accurate valuations and global coordination, which are often unavailable in online LLM serving. We formalize this tension as the \emph{Token Economics Impossibility Triangle}.

This paper makes four contributions:
\begin{enumerate}[leftmargin=*]
    \item It identifies computational token economics as a distinct research problem at the intersection of token economics, LLM inference systems, and computational resource allocation.
    \item It introduces the Token Economics Impossibility Triangle to characterize the trade-off among token-level granularity, real-time execution, and allocation optimality.
    \item It organizes the main technical challenges into real-time value accounting, constrained token allocation, and economic-aware system architecture.
    \item It outlines a research agenda for approximate token valuation, online allocation under uncertainty, transparent token accounting, verifiable inference, and hardware-aware economic mechanisms.
\end{enumerate}

The remainder of this paper is organized as follows. Section~II reviews background on token economics and LLM serving systems. Section~III introduces the Token Economics Impossibility Triangle. Sections~IV--VI analyze the three primary computational challenges: real-time value accounting, constrained allocation, and economic-aware architecture. Section~VII discusses broader implications and connections to adjacent fields. Section~VIII presents open problems and concludes the paper.

\section{Background and Related Work}

Large language models have made tokens a central unit of both computation and economic exchange. 
In early neural language modeling, tokens were mainly treated as linguistic symbols or as implementation artifacts of tokenization. 
In contemporary LLM services, however, tokens determine not only model input and output length, but also pricing, latency, KV-cache consumption, batching behavior, accelerator utilization, and the allocation of inference-time computation. 
This shift motivates a broader view in which tokens are economic objects whose value, cost, and allocation must be studied together with the computational systems that produce them.

This section reviews the literature relevant to this view. 
We first discuss the emergence of token economics and the interpretation of tokens as economic primitives. 
We then examine recent LLM serving systems, KV-cache management, speculative decoding, long-context inference, and test-time reasoning. 
These lines of work show that token-level decisions are already embedded in modern inference infrastructure, but are usually optimized through system-level heuristics rather than explicit economic mechanisms. 
This gap motivates the computational token economics perspective developed in this paper.

\subsection{Token Economics and the Economic Status of Tokens}

Recent work has begun to conceptualize tokens as economic primitives rather than merely technical units. 
Brass characterizes inference tokens as fundamental digital commodities and develops an inference-economics framework for analyzing production, consumption, pricing, platform structure, and infrastructure markets in AI services~\cite{brass_digital_lab_inference_2026}. 
Chen et al.\ further position tokens as crucial economic primitives in agentic AI systems, emphasizing their roles across individual agents, multi-agent interactions, ecosystem-level coordination, and security-sensitive environments~\cite{chen2026token}. 
Zhu argues that a single agentic request may pass through multiple economic layers, including routing, agent policy, serving infrastructure, and training pipelines, each of which implicitly compares marginal benefit, computational cost, latency cost, and risk cost~\cite{zhu2026agentic}. 
Xing similarly suggests that tokens are evolving from service outputs into infrastructure-level raw materials comparable to electricity, bandwidth, or cloud compute units~\cite{xing2026token}.

The central insight of this literature is that tokens are heterogeneous economic objects. 
Different tokens may have different marginal values, different production costs, and different effects on downstream utility. 
An instruction token can shape the behavior of an entire response; a retrieved document token may improve factuality only for some queries; a redundant context token may consume KV-cache memory while contributing little marginal value; and hidden reasoning tokens may affect final answer quality without being directly visible to users. 
This heterogeneity motivates token-level valuation, pricing, accounting, and allocation.

However, existing token-economics discussions often abstract away from the computational difficulty of implementing such mechanisms. 
Economic models may assume that valuations are available, that allocation objectives can be optimized, or that market-like mechanisms can clear at the relevant timescale. 
Production LLM inference systems operate in a different regime. 
Decisions about batching, prefill--decode scheduling, model routing, KV-cache placement, speculative execution, and admission control must be made under strict latency and memory constraints, often before the future value of a token is known. 
The key question is therefore not only how tokens should be valued in principle, but how token-economic mechanisms can be executed under the computational constraints of real inference systems.

\subsection{LLM Serving Systems and the Cost Structure of Tokens}

The computational cost of a token is shaped by the serving architecture that produces it. 
Modern LLM serving systems optimize throughput, latency, and memory efficiency through batching, scheduling, KV-cache management, and kernel-level acceleration. 
Although these systems are not usually framed as economic mechanisms, they already implement implicit token-allocation policies.

A representative example is vLLM, which introduces PagedAttention to manage KV cache using a virtual-memory-like paging mechanism~\cite{kwon2023vllm}. 
By reducing memory fragmentation and enabling flexible cache allocation, vLLM shows that KV-cache management is a first-order determinant of inference throughput. 
Sarathi-Serve improves serving efficiency by piggybacking decode operations with chunked prefills, thereby reducing scheduling inefficiencies between the prefill and decode phases~\cite{agrawal2024sarathi}. 
DistServe further separates prefill and decoding to improve goodput under heterogeneous workloads and latency constraints~\cite{zhong2024distserve}. 
SGLang provides a runtime for structured language-model programs and exploits prefix sharing, caching, and efficient execution of compound LLM workflows~\cite{zheng2024sglang}. 
Together, these systems indicate that token cost is not a fixed scalar. 
It depends on phase structure, sequence length, batching state, cache residency, memory bandwidth, and the interaction between user requests.

Kernel-level attention optimizations reinforce this point. 
FlashAttention-2 improves attention performance by reducing memory traffic and improving parallelism and work partitioning~\cite{dao2024flashattention2}. 
FlashAttention-3 further explores asynchronous execution and low-precision computation on modern accelerators~\cite{shah2024flashattention3}. 
Such work changes the marginal cost of processing long contexts and generating tokens. 
From an economic perspective, this means that the cost of a token is hardware- and architecture-dependent: the same token may have different effective costs depending on attention implementation, accelerator generation, batch composition, and memory layout. 
Therefore, token pricing or token allocation mechanisms cannot be designed independently of the serving system.

\subsection{KV-Cache Management and Approximate Token Valuation}

KV-cache management is one of the clearest places where system optimization approaches token valuation. 
During autoregressive decoding, each retained context token occupies KV-cache capacity and may influence future generation. 
When memory is scarce, the system must decide which tokens or KV states should be retained, compressed, or evicted. 
This is effectively an online allocation problem over tokens under a memory constraint.

Several recent methods exploit the observation that not all tokens contribute equally to future generation. 
H$_2$O identifies heavy-hitter tokens and uses them to guide KV-cache eviction for efficient generative inference~\cite{zhang2023h2o}. 
Scissorhands exploits the persistence of token importance during decoding and removes less important KV states at test time~\cite{liu2023scissorhands}. 
StreamingLLM shows that attention sinks enable stable streaming inference beyond the training context length, highlighting the special role of a small set of persistent tokens~\cite{xiao2024streamingllm}. 
SnapKV selects clustered important KV states before generation, reducing cache usage while preserving long-context performance~\cite{li2024snapkv}. 
PyramidKV further studies dynamic KV-cache compression through layer-wise and position-wise information funneling~\cite{cai2024pyramidkv}. 
More recently, TokenButler argues that token importance is predictable and uses lightweight prediction to support more efficient long-context inference~\cite{abdelfattah2025tokenbutler}.

These methods provide system-level evidence for the central assumption of token economics: tokens have unequal marginal value. 
However, they also reveal a computational difficulty. 
Exact token valuation would require counterfactual evaluation of how the output distribution or task utility changes when a token is retained, removed, compressed, or replaced. 
Such counterfactual valuation is usually too expensive for online inference. 
Existing systems therefore rely on approximate proxies such as attention patterns, persistence, clustering, or learned importance predictors. 
This illustrates a broader theme of computational token economics: economically meaningful valuation must often be approximated, amortized, or embedded into the serving architecture.

\subsection{Speculative Decoding and Approximate Token Production}

Autoregressive generation makes output tokens sequentially expensive. 
Speculative decoding and related acceleration methods reduce this cost by producing candidate tokens with cheaper mechanisms and verifying them against a target model. 
Leviathan et al.\ formalize speculative decoding as a method for accelerating transformer inference while preserving the distribution of the target model~\cite{leviathan2023speculative}. 
SpecInfer extends this idea with speculative inference trees and token-tree verification for high-throughput LLM serving~\cite{miao2024specinfer}. 
Medusa attaches multiple decoding heads to predict several future tokens in parallel, reducing reliance on a separate draft model~\cite{cai2024medusa}. 
EAGLE improves speculative sampling by rethinking feature uncertainty and using feature-level extrapolation~\cite{li2024eagle}. 
Lookahead decoding similarly accelerates generation by parallelizing candidate continuation and verification~\cite{fu2024lookahead}.

These methods can be interpreted as approximate token-production mechanisms. 
They trade additional computation, memory, and verification overhead for fewer expensive target-model decoding steps. 
The relevant optimization problem is therefore economic in structure: a system must decide whether producing and verifying speculative tokens is worth the expected reduction in latency and target-model computation. 
The answer depends on acceptance rate, model size, hardware utilization, batch state, and quality requirements. 
Thus, even when the goal is framed as inference acceleration, the underlying decision resembles an online cost--benefit calculation over candidate tokens.

Model routing systems expose a related issue at a coarser granularity. 
FrugalGPT studies cascaded use of LLM APIs to reduce cost while maintaining performance~\cite{chen2023frugalgpt}. 
RouteLLM learns to route queries between stronger and weaker models using preference data, balancing quality and cost~\cite{ong2024routellm}. 
Although these methods usually operate at the request level, they suggest a natural extension to token-level or segment-level routing in agentic and reasoning-intensive systems. 
The challenge is to determine how fine-grained routing can become before the overhead of valuation, routing, and verification outweighs the saved computation.

\subsection{Long-Context Inference and Token Relevance}

Long-context models make token allocation even more important. 
A larger context window increases the number of tokens that can be supplied to the model, but it does not guarantee that all tokens are equally useful or effectively used. 
Liu et al.\ show that language models often struggle to use information placed in the middle of long contexts, a phenomenon known as ``lost in the middle''~\cite{liu2024lost}. 
LongBench provides a bilingual multitask benchmark for long-context understanding and reveals substantial variation in long-context capabilities across models and tasks~\cite{bai2024longbench}. 
RULER further evaluates the effective context size of long-context models and shows that nominal context length may overstate usable context length~\cite{hsieh2024ruler}.

These findings have direct economic implications. 
Context-window capacity and KV-cache memory are scarce resources, and adding more tokens increases computational cost. 
If only a subset of context tokens contributes meaningfully to task utility, then context selection, prompt compression, retrieval filtering, and KV-cache eviction are all forms of constrained token allocation. 
Current systems often use heuristic proxies for relevance, such as recency, attention weight, retrieval score, or learned salience. 
A computational token-economics framework would ask a stronger question: how should the system allocate context and cache budget to maximize expected utility under latency, memory, and uncertainty constraints?

\subsection{Reasoning-Time Computation and Agentic Token Allocation}

Reasoning-oriented models and agentic workflows extend token allocation beyond visible input and output. 
Intermediate reasoning, search, verification, tool use, and inter-agent communication all consume tokens and system resources. 
Self-consistency demonstrates that sampling multiple reasoning paths can improve reasoning accuracy~\cite{wang2023selfconsistency}. 
Tree of Thoughts frames reasoning as deliberate search over intermediate thoughts rather than as a single left-to-right chain~\cite{yao2023tree}. 
Recent work on test-time scaling further studies how additional inference-time computation can improve performance and how such computation should be allocated~\cite{snell2024scaling}. 
These results suggest that output quality may depend critically on how many tokens are allocated to intermediate reasoning and how those tokens are organized.

Agentic systems make the allocation problem more complex. 
Toolformer shows that language models can learn to use external tools through self-supervised data generation~\cite{schick2023toolformer}. 
Voyager demonstrates an open-ended embodied agent that acquires reusable skills through interaction and code generation~\cite{wang2024voyager}. 
SWE-agent shows that LLM agents can solve real software engineering tasks by interacting with repositories through an agent-computer interface~\cite{yang2024sweagent}. 
In these systems, tokens are spent on planning, tool calls, memory retrieval, observation processing, code generation, reflection, and error correction. 
Their value is trajectory-dependent: a token may be useful not because it appears in the final response, but because it enables a later tool call, prevents an error, or improves the agent's plan.

This creates a difficult credit-assignment problem. 
In principle, cooperative game-theoretic tools such as marginal contribution or Shapley-style attribution could assign value to tokens, messages, agents, or intermediate actions. 
In practice, exact attribution requires repeated counterfactual evaluation and is usually infeasible during online inference. 
Consequently, agentic token allocation requires approximate, online, and uncertainty-aware mechanisms. 
This is one of the central reasons why token economics must be studied as a computational problem rather than only as a conceptual or market-design problem.

\subsection{Gap Between Economic Theory and System Implementation}

The literature reviewed above reveals a structural gap between token-economic theory and LLM system implementation. 
Token-economics work emphasizes pricing, valuation, exchange, and incentives, but often assumes away the difficulty of computing values and executing mechanisms at inference time. 
LLM systems work, in contrast, has developed sophisticated techniques for serving, scheduling, caching, speculative decoding, long-context inference, and model routing, but these techniques are generally optimized for throughput, latency, or memory rather than explicit economic objectives.

This gap has three dimensions. 
First, there is a granularity gap. 
Economic reasoning may call for token-level valuation, but production systems often make decisions at the request, batch, sequence, or cache-block level. 
Second, there is a timescale gap. 
Economic mechanisms may require estimation, optimization, and coordination, whereas inference systems must make decisions within strict latency budgets. 
Third, there is an architecture gap. 
Economic models often treat computation as an abstract cost, while real LLM systems are shaped by GPU memory hierarchy, KV-cache layout, attention kernels, batching behavior, and prefill--decode interactions.

This paper addresses these gaps by proposing \emph{Computational Token Economics} as a distinct research perspective. 
The objective is not merely to ask how tokens should be priced or valued in an abstract market, but to understand how token-economic mechanisms can be implemented under the constraints of real AI infrastructure. 
The following sections formalize this perspective through the Token Economics Impossibility Triangle, which captures the trade-off among fine-grained token valuation, real-time execution, and allocation optimality.

\section{Theoretical Framework: The Token Economics Impossibility Triangle}
\label{sec:impossibility_triangle}

\subsection{Motivation}

The preceding discussion suggests that tokens in modern LLM systems are no longer merely
linguistic units. They are also units of computation, memory consumption, latency, pricing,
and sometimes hidden reasoning effort. Once tokens are treated as economic primitives, a natural
question arises: how should a serving system value and allocate tokenized computation under
resource constraints?

At first sight, this question resembles a standard resource-allocation problem. Given a set of
tokens, requests, agents, or intermediate reasoning steps, one may attempt to assign each unit of
computation to the location where it yields the highest marginal utility. However, production LLM
serving differs sharply from offline economic optimization. Decisions about batching, routing,
prefill--decode scheduling, KV-cache placement, speculative execution, and admission control must
often be made at millisecond or sub-millisecond timescales. Meanwhile, the value of a token may
depend on future model states, downstream user utility, hidden reasoning trajectories, tool calls,
or verification outcomes, most of which are not fully observable at decision time.

This motivates a unifying theoretical lens. We argue that the primary difficulty of computational token economics is governed by a \textbf{trilemma}---a three-way tension among:
\emph{fine-grained economic accounting}, \emph{real-time execution}, and \emph{allocation
optimality}. Informally, the central design problem can be stated as follows:

\begin{quote}
\emph{Under strict latency, memory, and accelerator-utilization constraints, how can an LLM
serving system make economically meaningful token-level decisions that are as close as possible
to the decisions that would have been made with perfect information and unlimited computation?}
\end{quote}

This question is not only about pricing. It also concerns context selection, KV-cache management,
reasoning-budget allocation, request routing, batching, retrieval, tool invocation, agent credit
assignment, and verifiable inference. The framework introduced below is intended to explain why
these tasks are difficult in a common language and to clarify the trade-offs that later sections
analyze in more concrete forms.

\subsection{Three Axes of the Triangle}

We define the \textbf{Token Economics Impossibility Triangle} as a conceptual framework capturing
the tension among three desirable properties of token-economic systems:

\begin{enumerate}[leftmargin=*]
    \item \textbf{Granularity.}
    Granularity refers to the resolution at which the system measures token value and makes
    allocation decisions. A coarse system may treat an entire request, sequence, or batch as a
    single economic unit. A fine-grained system may attempt to assign marginal values to individual
    prompt tokens, generated tokens, reasoning tokens, retrieved chunks, or agent actions.
    Fine granularity is desirable because different tokens can have sharply different marginal
    effects on utility and cost. For example, an instruction token may determine the behavior of
    an entire response, whereas a redundant context token may consume KV-cache capacity with
    little benefit.

    \item \textbf{Real-time performance.}
    Real-time performance refers to the ability to execute economic decisions within the latency
    budget of online inference. In LLM serving, the relevant timescale is not a single number:
    routing and admission control may operate at request timescales, prefill scheduling at
    batch or sequence timescales, and decode-time decisions at token-generation timescales.
    Modern serving systems such as Orca, vLLM, and related engines demonstrate that throughput
    and latency depend critically on continuous batching, memory management, and careful
    scheduling~\cite{yu2022orca,kwon2023vllm,zhong2024distserve}. Any economic mechanism that
    adds substantial per-token overhead may therefore reduce the very utility it aims to optimize.

    \item \textbf{Optimality.}
    Optimality refers to the quality of economic decisions relative to an ideal allocation rule.
    In the strongest form, a system would compute the globally optimal allocation under complete
    information about token values, future requests, memory pressure, accelerator availability,
    and user utility. In practice, such information is unavailable or expensive to obtain. The
    system must therefore rely on approximate value estimates, online policies, amortized
    computation, or heuristics. Approximate optimality may still be useful if the loss can be
    bounded theoretically, estimated empirically, or controlled through monitoring.
\end{enumerate}

These three properties are individually desirable but jointly difficult to achieve. Fine-grained
valuation increases the number of economic objects and the cost of attribution. Real-time serving
limits the amount of sensing, counterfactual evaluation, and optimization that can be performed
per decision. Strong optimality often requires accurate valuations and global coordination, which
conflict with the locality and uncertainty of online inference. To make these tensions analytically tractable, the next subsection introduces a stylized online allocation model and defines measurable counterparts for the three axes of the trilemma.

\begin{figure}[t]
\centering
\includegraphics[width=0.8\textwidth]{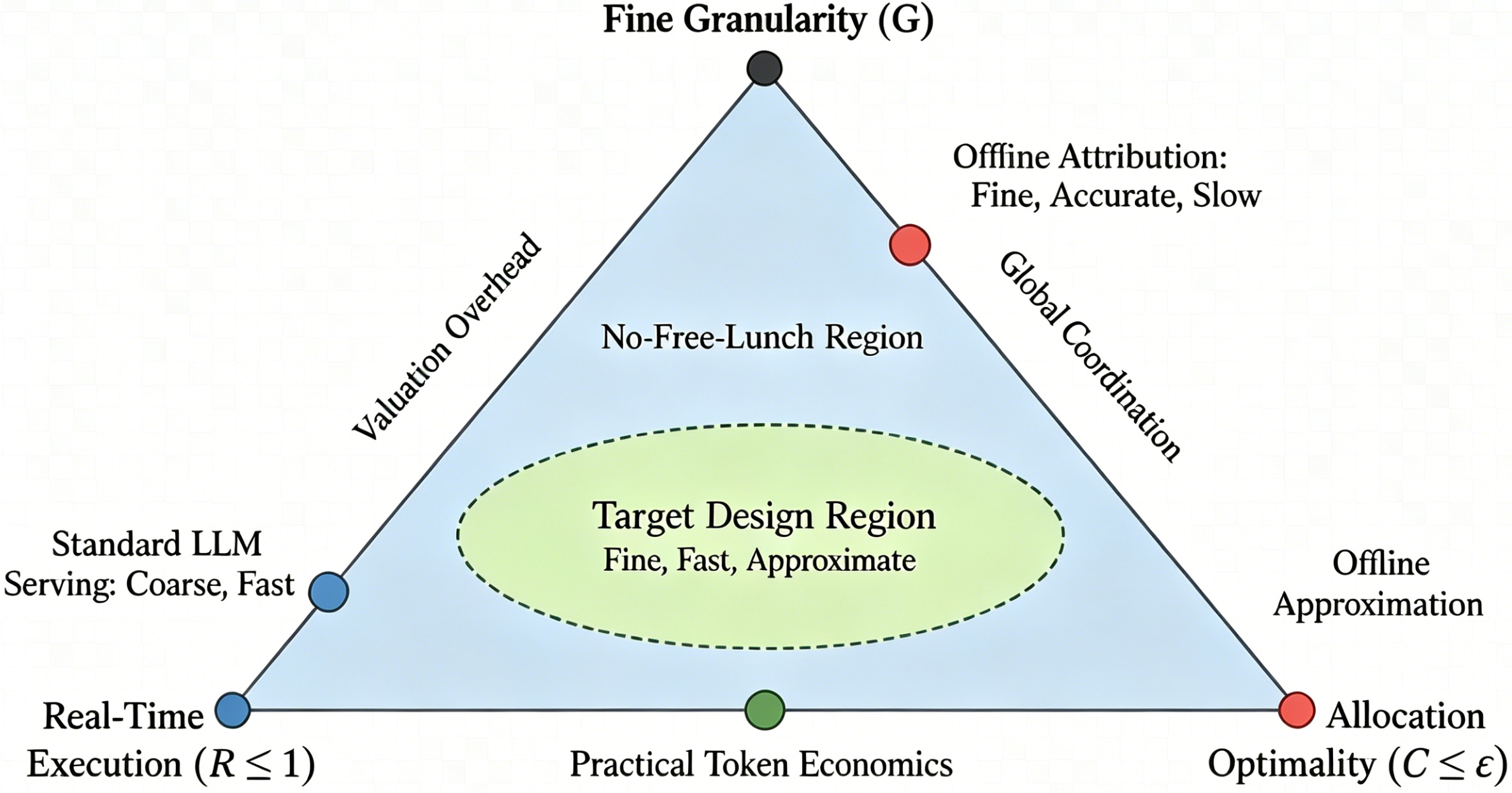}
\caption{The Token Economics Impossibility Triangle. Token-economic systems face a
three-way tension among fine-grained valuation, real-time execution, and allocation optimality.
Practical systems typically move toward the interior by relaxing at least one property through
coarsening, approximation, amortization, or hardware-aware design.}
\label{fig:token_economics_triangle}
\end{figure}

\subsection{A Formalized View}

We first introduce measurable counterparts for the three axes of the trilemma. 
Let $G(\phi_\theta, \mathcal{I}_t) = |\mathcal{I}_t|$ denote \textbf{granularity}, measured by the number of decision units (tokens, blocks, or spans) evaluated at step $t$. 
Let $R(\pi, \tau_t) = \big[T_{\mathrm{value}}(\phi_\theta) + T_{\mathrm{alloc}}(a_t)\big] / \tau_t$ denote the \textbf{real-time ratio}, i.e., the fraction of the per-step latency budget consumed by economic decision-making. 
Finally, let $O(\pi) = \mathbb{E}_t[\Delta_t]$ denote \textbf{optimality regret}, the expected economic loss relative to the offline benchmark defined below. 
The trilemma can then be restated as: under fixed serving constraints, no policy $\pi$ can simultaneously achieve $G \geq G^*$, $R \leq 1$, and $O \leq \epsilon$ without relaxing at least one axis through approximation, coarsening, or amortization.

Consider an online LLM serving system that receives a stream of requests
$\mathcal{R}=\{r_1,r_2,\ldots\}$. Each request induces a set of token-related decision units
$\mathcal{I}_t$ at time $t$, which may include prompt tokens, generated tokens, retrieved chunks,
reasoning steps, tool calls, or KV-cache blocks. The system chooses an allocation action
$a_t \in \mathcal{A}_t$, such as admitting a request, pruning context, assigning a request to a
model instance, allocating a reasoning budget, evicting cache blocks, or scheduling prefill and
decode work.

Let $v_i$ denote the unknown marginal economic value of unit $i \in \mathcal{I}_t$, and let
$c_i$ denote its resource cost, including compute, memory bandwidth, KV-cache footprint, and
latency impact. A stylized offline allocation problem can be written as
\begin{equation}
    \max_{a \in \mathcal{A}} \quad U(a; \mathbf{v}) - C(a; \mathbf{c})
    \quad
    \text{s.t.} \quad
    M(a) \leq B_M,\;\; L(a) \leq B_L,\;\; H(a) \leq B_H,
    \label{eq:offline_allocation}
\end{equation}
where $U$ is the utility induced by the allocation, $C$ is the resource cost, and
$B_M$, $B_L$, and $B_H$ denote memory, latency, and hardware-capacity budgets. If the values
$\mathbf{v}$ were known and the optimization could be solved offline, Eq.~\eqref{eq:offline_allocation}
would define an ideal benchmark.

In online inference, however, the system observes only partial information. It must construct
a value estimate $\hat{v}_i$ under a computation budget $\tau_t$:
\begin{equation}
    \hat{\mathbf{v}}_t = \phi_\theta(x_t, s_t; \tau_t),
    \qquad
    T_{\mathrm{value}}(\phi_\theta) + T_{\mathrm{alloc}}(a_t) \leq \tau_t,
    \label{eq:value_budget}
\end{equation}
where $x_t$ is the current input, $s_t$ is the serving-system state, and
$\phi_\theta$ is a value-estimation mechanism. The budget $\tau_t$ may be extremely small during
decode-time scheduling. The realized decision quality depends on both the value-estimation error
and the optimization error:
\begin{equation}
    \Delta_t
    =
    \underbrace{
    \left[
    U(a_t^\star;\mathbf{v}) - C(a_t^\star;\mathbf{c})
    \right]
    -
    \left[
    U(a_t;\mathbf{v}) - C(a_t;\mathbf{c})
    \right]
    }_{\text{economic regret relative to the offline benchmark}},
    \label{eq:regret}
\end{equation}
where $a_t^\star$ denotes the ideal offline action. The objective of computational token economics
is not necessarily to eliminate $\Delta_t$, which is generally unrealistic, but to reduce it under
strict serving constraints.

This formulation exposes the three axes of the triangle. Increasing granularity enlarges
$\mathcal{I}_t$ and often increases the cost of estimating $\hat{\mathbf{v}}_t$. Improving optimality
requires reducing estimation and optimization errors, often by using more counterfactual
evaluation, global coordination, or combinatorial optimization. Maintaining real-time performance
requires bounding $T_{\mathrm{value}}$ and $T_{\mathrm{alloc}}$, which forces approximation,
coarsening, or amortization. Thus, the three measurable quantities $(G, R, O)$ are jointly constrained under any fixed serving budget. Pushing the frontier on one axis without relaxing the others requires architectural innovation---for example, hardware support for faster value estimation (reducing $R$ without coarsening $G$), or amortized attribution that shifts computation offline (reducing $R$ without increasing $O$). The remainder of this paper analyzes these relaxations in concrete technical settings.

\begin{figure}[t]
\centering
\includegraphics[width=0.8\textwidth]{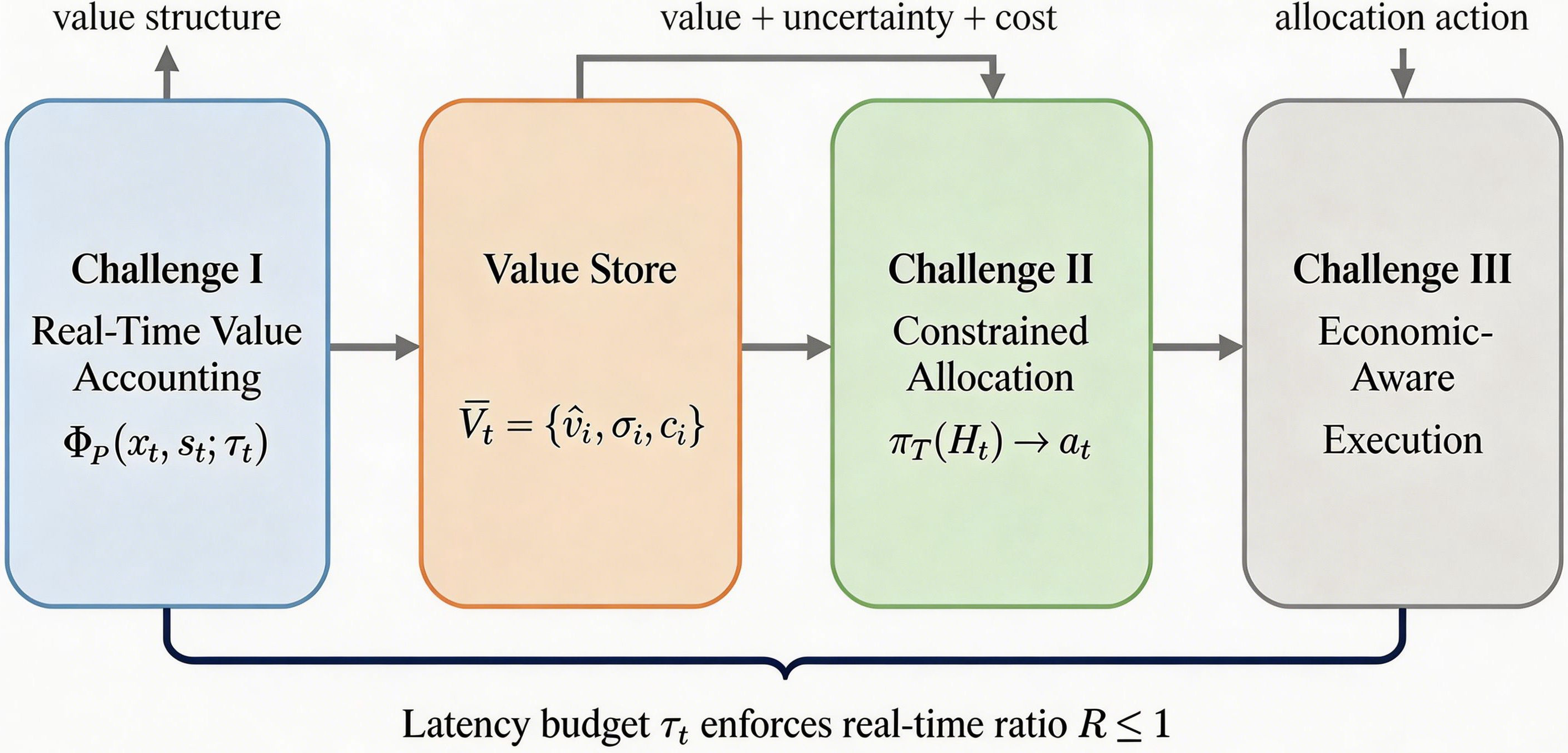}
\caption{Interface between the three crucial challenges. Challenge~I (Sensing) produces a value structure $\hat{\mathcal{V}}_t$ under budget $\tau_t$. Challenge~II (Allocation) consumes this structure to select an action $a_t$. Challenge~III (Architecture) executes the action within the serving stack. The latency budget $\tau_t$ couples all three stages and enforces the trilemma's real-time axis.}
\label{fig:challenge_interface}
\end{figure}

\subsection{A Conditional No-Free-Lunch Statement}

We do not claim a universal information-theoretic impossibility theorem 
in the sense of, e.g., the CAP theorem or Arrow's theorem.  
Rather, the triangle should be understood as a \emph{conditional} 
no-free-lunch principle for online token-economic mechanisms.

\begin{proposition}[Token Economics Trilemma, Informal]
\label{prop:trilemma}
Consider an online LLM serving system operating under the following 
conditions:
\begin{enumerate}[label=(\roman*)]
    \item \textbf{Heterogeneous and a priori unknown values.} 
    Token-level marginal values $v_i$ vary across instances and are not 
    fully observable before the allocation decision is made.
    \item \textbf{Non-negligible online sensing.} 
    Obtaining a sufficiently accurate value estimate for a token 
    requires non-zero online computation---e.g., counterfactual probing, 
    verifier calls, or dynamic predictor inference---whose residual 
    cost cannot be entirely eliminated by offline pre-training or 
    amortized attribution alone.
    \item \textbf{Strict latency budget.} 
    The total online computation budget per decision step is bounded 
    by $\tau$, with $\tau \ll N$ relative to the per-token sensing cost.
    \item \textbf{Combinatorial allocation.} 
    The action space includes subset-selection over $N$ active tokens 
    under a budget $B<N$.
\end{enumerate}
Under (i)--(iv), no online policy can simultaneously achieve
\begin{itemize}
    \item \textbf{Granularity:} $G=N$ (token-level decisions),
    \item \textbf{Real-time:} $R\le 1$ (within budget $\tau$),
    \item \textbf{Optimality:} $O=o(1)$ (sub-constant regret relative 
    to the offline optimum with full information).
\end{itemize}
At least one of the following relaxations is necessary: coarser 
granularity, approximate valuation, approximate allocation, 
amortized/offline coordination, or hardware-assisted sensing.
\end{proposition}

\paragraph{Proof intuition.} 
Conditions (ii)--(iii) imply that any online policy can explicitly 
evaluate at most $k=\lfloor \tau/c\rfloor<N$ tokens (for some constant 
per-token sensing cost $c$) before committing to a subset 
$S\subseteq[N]$ with $|S|\le B$. Consider an adversarial family in 
which exactly $B$ tokens have $v_i=1$ and the remaining $N-B$ have 
$v_i=0$, with the high-value set chosen uniformly at random. 

Among the $k$ queried tokens, the expected number of high-value 
tokens found is $kB/N$. For the $N-k$ unqueried tokens, the policy 
can retain at most $B$ of them, and each unqueried token is a 
high-value token with marginal probability $B/N$ by symmetry. Hence 
the expected contribution from unqueried tokens is at most $B^2/N$. 
The total expected retained value is therefore bounded by
\[
\mathbb{E}[V(S)] \;\le\; \frac{kB}{N} + \frac{B^2}{N}
\;=\; \frac{(k+B)B}{N}.
\]
Since $k\ll N$, the expected regret against the offline optimum 
(which knows all $B$ high-value tokens) satisfies
\[
\Delta_t \;=\; B - \mathbb{E}[V(S)]
\;\ge\; B\Bigl(1-\frac{k+B}{N}\Bigr)
\;=\; \Omega(1).
\]
Thus, under fixed $\tau$ (hence fixed $k$), exact token-level 
valuation and hard real-time execution are incompatible with 
sub-constant regret unless $N$ is bounded---which contradicts the 
long-context scaling regime. The only escapes are to coarsen 
granularity (reducing effective $N$), replace exact valuation with 
a lightweight proxy (bypassing the per-token cost), or accept 
approximate optimality.

\subsection{Design Regimes Induced by the Triangle}

The triangle induces several recurring design regimes, summarized in
Table~\ref{tab:triangle_design_regimes}. Each regime relaxes a different vertex.

\begin{table}[t]
\centering
\caption{Representative design regimes under the Token Economics Impossibility Triangle.}
\label{tab:triangle_design_regimes}
\begin{tabular}{@{}p{3.1cm}p{2.3cm}p{2.3cm}p{2.6cm}p{4.0cm}@{}}
\toprule
\textbf{Design regime} & \textbf{Granularity} & \textbf{Real-time} & \textbf{Optimality} & \textbf{Typical mechanism} \\
\midrule
Conventional serving &
Request/batch level &
Strong &
Heuristic &
Continuous batching, routing, cache management, throughput-oriented scheduling \\

Offline economic analysis &
Token or step level &
Weak &
Strong or interpretable &
Counterfactual evaluation, attribution, exhaustive profiling, offline replay \\

Naive token-economic optimization &
Token level &
Weak &
Potentially strong &
Solving a large optimization problem per request or per decoding stage \\

Practical computational token economics &
Token or small-block level &
Strong &
Approximate &
Learned value proxies, amortized attribution, online approximation, hardware-aware scheduling \\

Auditable or verifiable serving &
Sequence/trace level &
Moderate &
Task-dependent &
Transparent accounting, proof or verification mechanisms, selective auditing \\
\bottomrule
\end{tabular}
\end{table}

The most promising region is not the unattainable corner where all three properties are maximized
simultaneously, but the interior region where systems make economically meaningful decisions at
the finest feasible granularity while preserving serving performance. This points to several
methodological directions:

\begin{itemize}[leftmargin=*]
    \item \textbf{Approximate value proxies.}
    Instead of computing exact marginal values, systems can learn lightweight predictors of token
    importance, context utility, or reasoning benefit. These predictors may be trained offline and
    invoked online with bounded overhead.

    \item \textbf{Amortized attribution.}
    Expensive attribution can be performed offline, periodically, or on sampled traces. The results
    can then be distilled into compact policies or value models used during inference.

    \item \textbf{Coarsened economic units.}
    Rather than valuing every token independently, systems may group tokens into semantic spans,
    retrieved chunks, KV-cache pages, reasoning blocks, or agent actions. This sacrifices some
    granularity but can greatly reduce optimization cost.

    \item \textbf{Online approximation and competitive policies.}
    Since future requests and future token values are uncertain, policies can be evaluated in terms
    of regret, competitive ratio, or empirical utility under workload distributions, drawing on
    online optimization and online algorithms~\cite{borodin1998online,hazan2016introduction}.

    \item \textbf{Hardware-aware economic mechanisms.}
    Economic decisions should be aware of accelerator constraints, memory hierarchy, batching
    behavior, and prefill--decode asymmetry. A token that appears valuable in an abstract economic
    model may be expensive or disruptive under a concrete serving architecture.
\end{itemize}

\subsection{Mapping the Three Crucial Challenges to the Triangle}

The three technical challenges analyzed in the following sections can be viewed as different
projections of the same triangle.

\paragraph{Challenge I: real-time value accounting.}
This challenge lies primarily on the edge between granularity and real-time performance. Token-level
economic accounting requires estimating the marginal contribution of individual tokens, context
segments, reasoning steps, or hidden computation. Exact accounting is often too expensive, while
coarse accounting may ignore economically important heterogeneity. The central question is how to
construct value estimates that are sufficiently informative, sufficiently cheap, and sufficiently
transparent for online use.

\paragraph{Challenge II: constrained token allocation.}
This challenge lies primarily on the edge between real-time performance and optimality. Given value
estimates and resource constraints, the system must allocate limited context length, decoding
budget, KV-cache capacity, accelerator time, retrieval calls, or verification effort. Many such
problems are combinatorial, online, and uncertain. The central question is how to design allocation
policies that are fast enough for serving while retaining theoretical or empirical guarantees.

\paragraph{Challenge III: economic-aware system architecture.}
This challenge concerns the full triangle. Even accurate value models and good allocation
algorithms may fail if they are not integrated into the serving stack. Economic mechanisms must
therefore be co-designed with batching, scheduling, memory management, model routing, speculative
decoding, and verification. The central question is how to build architectures in which economic
signals become first-class scheduling and accounting signals rather than offline annotations.

\subsection{Relation to Classical Impossibility and Trade-off Results}

The proposed triangle is related to, but distinct from, several classical trade-off frameworks.
In distributed systems, the CAP theorem formalizes a tension among consistency, availability, and
partition tolerance~\cite{gilbert2002brewer}. In mechanism design, impossibility and trade-off
results show that incentive compatibility, efficiency, budget balance, and individual rationality
cannot always be achieved simultaneously~\cite{myerson1983efficient}. In online algorithms and
online convex optimization, decisions must be made before future inputs are known, leading to
competitive-analysis and regret-based formulations~\cite{borodin1998online,hazan2016introduction}.

The Token Economics Impossibility Triangle differs in its object of analysis. It concerns the
computational feasibility of economic mechanisms inside probabilistic, high-throughput AI
inference systems. The relevant scarcity is not only monetary or informational, but also
architectural: memory bandwidth, KV-cache capacity, accelerator occupancy, batching efficiency,
and latency budgets directly shape the space of feasible economic decisions. This makes
computational token economics a hybrid problem at the intersection of economic theory, online
optimization, and AI systems design.

\paragraph{Orthogonal dimensions.} 
The triangle captures computational feasibility. 
Other economically significant tensions---notably 
\emph{fairness} across tenants, \emph{incentive compatibility} 
against strategic manipulation, \emph{privacy budgets} 
(e.g., differentially-private KV-cache accounting), and 
\emph{verifiable auditability}---are largely orthogonal to the 
computational axes. They impose additional constraints that 
interact with $G$, $R$, and $O$, but their full treatment lies 
outside the scope of this paper; we touch upon them in 
\S\ref{sec:broader} and leave systematic integration to future work.

\section{Challenge I: Real-Time Token Value Accounting (Sensing)}

\subsection{The Economic Problem}

\textbf{Key question.} How much is a token worth at the moment when an inference system must make a decision about it?

In computational token economics, tokens are not homogeneous units of text. They are heterogeneous economic objects with different marginal values, production costs, and effects on downstream utility. A system-prompt token may constrain the behavior of an entire interaction; a user-query token may determine the task objective; a retrieved-context token may or may not affect the final answer; an intermediate reasoning token may improve solution quality while remaining invisible to the user; and an output token incurs sequential autoregressive decoding cost. Recent work on inference economics and token economics has therefore argued that tokens should be treated not only as linguistic or computational artifacts, but also as pricing units, production factors, and allocation objects in AI service markets~\cite{chen2026token,zhu2026agentic,zhong2025token,smolin2025economics}.

The key economic object is not the raw token count, but the token's \emph{marginal contribution} to a service objective under resource constraints. A useful abstraction is to define the net value of token $i$ under serving state $s$ as
\begin{equation}
    v_i(s)
    =
    \mathbb{E}\!\left[
        U(Y \mid s)
        -
        U(Y \mid s^{-i})
        \,\middle|\, s
    \right]
    -
    \lambda
    \mathbb{E}\!\left[
        C_i(s)
        \,\middle|\, s
    \right],
    \label{eq:token_marginal_value}
\end{equation}
where $U(Y \mid s)$ is the utility of the eventual service outcome $Y$ under state $s$, $s^{-i}$ denotes a counterfactual serving state in which token $i$ is removed, masked, compressed, evicted, or otherwise not allocated the same resources, $C_i(s)$ is the resource cost associated with the token, and $\lambda$ converts resource cost into the same unit as utility. This definition is deliberately state-dependent. The same token can have different economic value depending on the prompt, model, cache state, latency budget, user objective, pricing rule, and current decoding trajectory.

\paragraph{Value versus price.} 
Throughout this paper, $v_i$ (or $v_i(s)$) denotes the 
\emph{internal marginal economic value}---the net contribution of 
token $i$ to service utility under resource constraints, as defined 
in Eq.~(4). We reserve $p_i$ for the \emph{external price} or 
charging rule applied to token $i$ by service contracts or markets. 
The distinction is essential: $v_i$ guides \emph{allocation}, while 
$p_i$ governs \emph{billing and incentives}. A token may carry high 
internal value ($v_i \gg 0$) yet low external price, or vice versa. 
For example, a system prompt that encodes safety constraints may 
have large $v_i$ for the serving system but be invisible to billing; 
conversely, peak-hour congestion pricing may set $p_i$ high for 
tokens whose $v_i$ is moderate. This separation is important 
because recent work on token pricing, tokenization incentives, and 
token futures highlights that pricing tokens purely by count can 
create distortions when tokens differ in utility, cost, or 
transparency~\cite{zhong2025token,artola2025overcharging,xing2026token}. 
We do not pursue the optimal design of $p_i$ here; our focus is 
on how $v_i$ can be estimated and acted upon under strict serving 
constraints. The relationship between $p_i$ and $v_i$---and the 
incentive distortions that arise when they diverge---is an 
important complement to the present framework.

\paragraph{Why it matters.} 
Real-time token value accounting is the sensing layer of computational 
token economics. Without some estimate of token value, a serving 
system cannot make economically informed decisions about which context 
tokens to retain, which KV-cache entries to compress or evict, how 
much reasoning budget to allocate, how to price hidden reasoning 
tokens, how to route requests across models, or how to assign credit 
across agents and tools. However, the estimate must often be produced 
before the final output and user utility are observed. This makes 
token value accounting an online estimation problem under uncertainty, 
rather than a purely offline attribution or pricing problem.

\subsection{The Computational Challenge}

\textbf{Requirement.} A deployable value-accounting mechanism must estimate token-level, span-level, or block-level value under the same constraints that govern real-time LLM inference:
\begin{itemize}[leftmargin=*]
    \item \textbf{Low latency.} The estimator must add negligible overhead relative to prefill and decode execution. A mechanism that requires repeated model calls or global optimization may be economically meaningful but infeasible for online serving.
    
    \item \textbf{Long-context scalability.} The estimator must remain feasible when the context contains many candidate tokens. This is increasingly important as long-context models expose large context windows, while benchmarks show that larger context length does not imply uniform or effective use of all tokens~\cite{reid2024gemini15,dubey2024llama3,hsieh2024ruler,liu2024lost,bai2024longbench}.
    
    \item \textbf{Causality.} Many serving decisions must be made before the final response is known. The estimator therefore cannot rely on future tokens, post-hoc human feedback, or outcome information unavailable at decision time.
    
    \item \textbf{Hardware awareness.} Token value estimates must be compatible with batching, prefill--decode scheduling, KV-cache placement, memory bandwidth, speculative decoding, and accelerator utilization. These constraints are central to modern LLM serving systems~\cite{yu2022orca,kwon2023vllm,agrawal2024sarathi,zhong2024distserve}.
    
    \item \textbf{Decision relevance.} The estimator need not recover a philosophically exact value. It must be accurate enough to improve downstream decisions such as cache eviction, compression, routing, pricing, admission control, or reasoning-budget allocation.
\end{itemize}

\textbf{Why it is hard.}

\textit{1. Exact marginal valuation requires counterfactual inference.}
The most direct way to estimate the value of token $i$ is to compare the model outcome with and without that token. If done naively, leave-one-out evaluation for a sequence of $n$ tokens requires $O(n)$ additional evaluations of the model or scoring function. This is already too expensive for long-context online inference. More principled cooperative-game-theoretic notions, such as the Shapley value, require averaging marginal contributions over all possible coalitions~\cite{shapley1953value}:
\begin{equation}
    \phi_i(F)
    =
    \sum_{S \subseteq N \setminus \{i\}}
    \frac{|S|!(n-|S|-1)!}{n!}
    \left[
        F(S \cup \{i\}) - F(S)
    \right],
    \label{eq:shapley_value}
\end{equation}
where $N$ is the set of $n$ tokens and $F(S)$ denotes the utility obtained when the subset $S$ is available. This definition is attractive because it gives a principled attribution rule. However, exact computation requires evaluating exponentially many subsets in the worst case. Recent token-level Shapley attribution methods therefore rely on approximation rather than exact online computation~\cite{xiao2025tokenshapley}.

\textit{2. Token value is interaction-dependent.}
A token's contribution is rarely independent of other tokens. An instruction token may only matter when combined with a later user query; a retrieved passage may only be valuable if it contains evidence relevant to the question; and an intermediate reasoning token may become valuable because later reasoning steps depend on it. Therefore, token value is generally non-additive:
\begin{equation}
    V(x_1,\ldots,x_n)
    \neq
    \sum_{i=1}^{n} v_i .
    \label{eq:non_additivity}
\end{equation}
Any scalar token score is therefore an approximation to a higher-order interaction structure. This is one reason why attention weights, likelihoods, or heuristic importance scores should not be interpreted as complete economic valuations.

\textit{3. Value is multi-dimensional.}
A token may contribute to factual accuracy, instruction following, safety, format compliance, reasoning depth, latency reduction, cache reuse, or user satisfaction. These objectives can conflict. Additional reasoning tokens may improve solution quality but increase latency and cost; safety-relevant tokens may have little visible semantic contribution to the final answer but high value in preventing harmful or noncompliant outputs. A scalar value function therefore requires an explicit aggregation rule, for example:
\begin{equation}
    U
    =
    \alpha_{\mathrm{acc}} U_{\mathrm{acc}}
    +
    \alpha_{\mathrm{safety}} U_{\mathrm{safety}}
    +
    \alpha_{\mathrm{format}} U_{\mathrm{format}}
    +
    \alpha_{\mathrm{user}} U_{\mathrm{user}}
    -
    \lambda_{\mathrm{lat}} C_{\mathrm{lat}}
    -
    \lambda_{\mathrm{mem}} C_{\mathrm{mem}}
    -
    \lambda_{\mathrm{comp}} C_{\mathrm{comp}} .
    \label{eq:multi_objective_utility}
\end{equation}
The weights in this objective encode deployment-specific preferences. Thus, there is no universal token-value function independent of the service objective.

\textit{4. Value changes over time and is only partially observable.}
The value of a token can evolve during decoding. A context token that initially appears irrelevant may later become important; a token that receives attention early may become redundant after the model resolves the user's intent; and hidden reasoning tokens may affect quality and cost without being exposed to the user. Recent work on counting invisible reasoning tokens in opaque commercial APIs illustrates that even measuring the amount of hidden tokenized computation can be difficult when the serving system is not transparent~\cite{sun2025coin}. Real-time value accounting must therefore estimate value under partial observability.

\textit{5. Sensing itself consumes resources.}
Value accounting is not free. Computing attributions, maintaining additional predictors, storing token scores, or invoking verifiers consumes compute, memory, and latency budget. This creates a recursive economic problem: the value of measuring token value must itself exceed the cost of measurement. This is the sensing-side manifestation of the broader impossibility triangle among granularity, real-time performance, and optimality.

\subsection{Existing Approaches and Limitations}

\textbf{Attention-based proxies.}
A lightweight approach is to use attention statistics as token-importance signals. Such proxies are attractive because attention is already computed during inference, especially in optimized attention implementations~\cite{dao2022flashattention,dao2024flashattention2,shah2024flashattention3}. Attention-based methods are also related to KV-cache compression and retention strategies that exploit the observation that not all tokens contribute equally to future generation~\cite{zhang2023h2o,liu2023scissorhands,xiao2024streamingllm,cai2024pyramidkv,li2024snapkv}.

However, attention is not economic value. A token can receive high attention because it is syntactically central, nearby, or useful for local coherence while contributing little to final task utility. Conversely, a safety constraint, system instruction, or early task-defining token may have high economic value even if its attention pattern is indirect. Attention statistics are therefore useful sensing signals, but they should be treated as proxies rather than complete value measures.

\textbf{Perturbation-, gradient-, and Shapley-style attribution.}
Perturbation methods estimate value by masking, deleting, replacing, or compressing tokens and measuring the change in output quality or model score. Gradient-based methods estimate sensitivity of a target objective with respect to token representations. Shapley-style methods provide a more principled marginal-contribution framework and have recently been applied to token-level context attribution~\cite{xiao2025tokenshapley}. Related Shapley-based ideas have also been used for credit assignment in multi-agent cooperation~\cite{hua2025shapley}.

The limitation is computational cost. These methods usually require additional forward passes, backward passes, or sampling over token subsets. They are therefore better suited for offline analysis, auditing, dataset valuation, or selective refinement than for assigning high-fidelity values to every token during every online inference request.

\textbf{Learned token-importance predictors.}
Another approach is to train a lightweight model that predicts token importance before or during inference. TokenButler shows that token importance can be predictable and that learned predictors can support more efficient long-context inference~\cite{abdelfattah2025tokenbutler}. This direction is promising because it amortizes expensive attribution into a cheaper online estimator.

Nevertheless, token importance is not the same as economic value. A learned predictor may be trained to preserve output quality, attention structure, or benchmark accuracy, while economic value additionally depends on latency, memory pressure, pricing, safety, and user utility. Moreover, a predictor trained under one model architecture, workload, or serving policy may not transfer reliably to another. Learned token-value models therefore require calibration to the specific economic decision they are intended to support.

\textbf{KV-cache compression and memory-aware selection.}
KV-cache methods such as H$_2$O, Scissorhands, StreamingLLM, PyramidKV, KIVI, and SnapKV show that context tokens can be treated differently from a memory-allocation perspective~\cite{zhang2023h2o,liu2023scissorhands,xiao2024streamingllm,cai2024pyramidkv,liu2024kivi,li2024snapkv}. These methods are highly relevant to token value accounting because they operationalize the idea that some tokens are more important to retain than others.

However, their objective is usually system efficiency or output preservation, not full economic valuation. A KV-cache policy may identify tokens that are important for generation continuity, but it may not account for user-level utility, pricing fairness, hidden reasoning cost, or incentive compatibility. Thus, KV-cache importance is an important component of token value, but not the entire concept.

\textbf{Utility-aware pricing and data valuation.}
Utility-aware data pricing connects token-level or data-level quality signals to downstream utility and empirical training gain~\cite{xu2026utility}. Such work is relevant because it moves beyond raw token counts and asks how informational units contribute to measurable utility. It also aligns with broader token-economic work that studies token allocation and optimal pricing~\cite{smolin2025economics,zhong2025token}.

However, training-data valuation and inference-time token accounting differ in important ways. Training-data valuation can often be performed offline and evaluated over datasets or training gains. Inference-time valuation must operate online, under request-specific objectives, strict latency budgets, and partial information. Therefore, utility-aware pricing methods provide useful conceptual tools, but they cannot be directly transplanted into real-time inference without adaptation.

\textbf{Tokenization transparency and hidden-token accounting.}
Token-level pricing creates incentives around tokenization, transparency, and billing. Artola Velasco et al.\ study how tokenization can affect transparency and incentives in LLM charging~\cite{artola2025overcharging}. Sun et al.\ examine the problem of counting invisible reasoning tokens in opaque commercial APIs~\cite{sun2025coin}. These issues are directly relevant to value accounting because users may be charged for tokens whose role, necessity, or even existence is not fully observable.

The limitation is that transparency does not by itself solve valuation. Counting tokens is easier than estimating their marginal contribution. A system may reveal how many input, output, or hidden reasoning tokens were used, but still not reveal whether those tokens were economically valuable.

\textbf{Verifiable inference and auditability.}
When token accounting affects payment, service guarantees, or trust, value-relevant events may need to be auditable. Verifiable inference mechanisms such as TopLoc point toward systems in which certain inference events can be checked without fully trusting the service provider~\cite{ong2025toploc}. Such mechanisms are not token-value estimators by themselves, but they may provide infrastructure for verifying claims about token usage, reasoning effort, or computation performed.

\begin{figure}[t]
\centering
\includegraphics[width=0.86\textwidth]{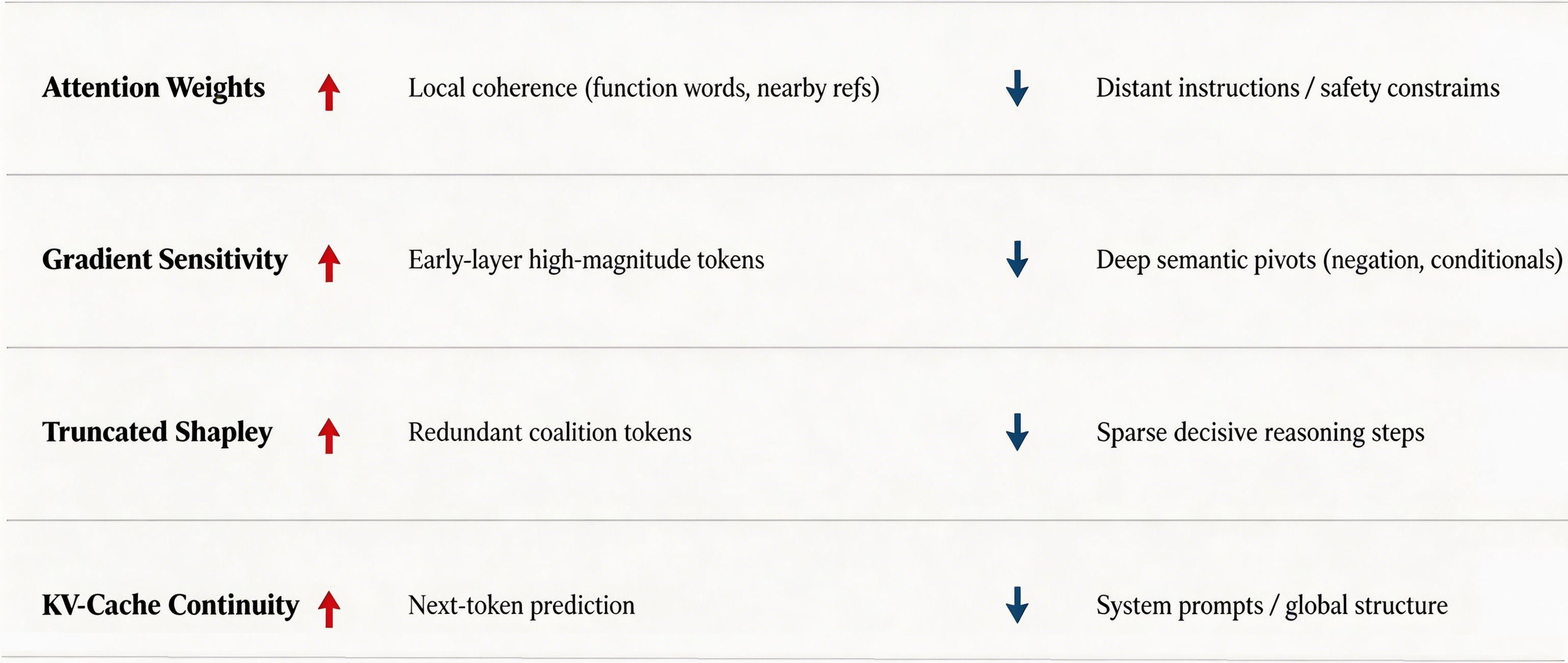}
\caption{Systematic directional biases of common value proxies relative to true economic value $v_i$. Each row shows one proxy family; red upward arrows indicate token types that are systematically \emph{overvalued}, while blue downward arrows indicate types that are systematically \emph{undervalued}.}
\label{fig:value_proxy_bias}
\end{figure}

\paragraph{Systematic biases of value proxies.}
The proxies reviewed above are not merely noisy estimates of economic value; they exhibit \emph{structured, directional biases} that can mislead allocation decisions. As illustrated in Figure~\ref{fig:value_proxy_bias}, attention-based proxies systematically overvalue tokens that contribute to local syntactic coherence (e.g., frequent function words) while undervaluing distant but critical instructions or safety constraints. Gradient-based proxies tend to overweight early-layer tokens with large activation magnitudes, while potentially missing deep semantic pivots such as negation or conditionals. Shapley-style approximations, when truncated or sampled, can overvalue redundant tokens that appear in many sampled coalitions and undervalue sparse but decisive reasoning steps. KV-cache compression proxies optimize for generation continuity, which overvalues tokens needed for next-token prediction while undervaluing tokens that shape global response structure (e.g., system prompts). Recognizing these directional biases is a prerequisite for \emph{decision calibration}: a proxy should be evaluated not by its correlation with some offline ground truth, but by the regret it induces in the specific downstream decision (eviction, pricing, routing) for which it is used.

\subsection{Open Problems}

\begin{enumerate}[leftmargin=*]
    \item \textbf{Decision-calibrated value proxies.}
    Can we design token-value proxies that are calibrated to concrete serving decisions, rather than merely correlated with attention, likelihood, or benchmark accuracy? As Figure~\ref{fig:value_proxy_bias} illustrates, existing proxies exhibit systematic directional biases: attention overweights local coherence, gradients favor early-layer magnitudes, and KV-cache proxies prioritize next-token continuity. The right value signal for KV-cache eviction may differ from the right signal for pricing, routing, or reasoning-budget allocation. A proxy should be evaluated by the \emph{decision regret} it induces, not by its correlation with an offline ground truth.

    \item \textbf{From sequence-level utility to token-level credit.}
    Given only sequence-level outcomes, such as answer correctness, verifier success, user satisfaction, or downstream task completion, how should value be decomposed into token-level or span-level contributions without full counterfactual evaluation?

    \item \textbf{Valuation under uncertainty.}
    Since future tokens and final outcomes are unknown at decision time, should a system optimize expected token value, risk-adjusted value, worst-case value, or value of information? How should this choice depend on the application domain and service-level objective?

    \item \textbf{Multi-objective aggregation.}
    How should semantic utility, safety utility, formatting utility, latency cost, memory cost, and monetary cost be aggregated? Is a scalar token-value score sufficient, or should systems maintain vector-valued token accounts and defer scalarization until a specific decision is made?

    \item \textbf{Hidden reasoning-token valuation.}
    Reasoning-oriented models may consume internal tokens that affect both quality and cost. How can such hidden reasoning tokens be counted, valued, priced, or audited without exposing proprietary model internals or sensitive reasoning traces~\cite{sun2025coin}?

    \item \textbf{Incentive compatibility.}
    If token value affects pricing or allocation, users and agents may strategically manipulate prompts, formatting, or tokenization to appear more valuable or to reduce charges. How can token-value accounting mechanisms remain robust to such behavior~\cite{artola2025overcharging}?

    \item \textbf{Cost of sensing.}
    When is it economically worthwhile to measure token value at all? A fine-grained estimator may improve allocation quality but consume additional compute and latency. Characterizing the optimal granularity of sensing remains an open problem.
\end{enumerate}

\subsection{Promising Directions}

\textbf{Hierarchical value accounting.}
Instead of estimating precise values for every token, systems can first compute coarse span-level or block-level scores and then refine only the candidates involved in high-impact decisions. For example, a system might use inexpensive attention or cache statistics to identify likely low-value context blocks, while reserving more expensive attribution for borderline eviction, pricing, or verification decisions. This approach explicitly trades granularity for real-time feasibility.

\textbf{Amortized attribution.}
Expensive attribution can be performed offline on sampled workloads, recurring prompts, tool specifications, or frequently retrieved documents. The resulting estimates can then be distilled into lightweight online predictors. This direction is especially suitable for static or semi-static tokens such as system prompts, instruction templates, tool descriptions, and common retrieval passages.

\textbf{Hardware-aware value functions.}
Token value should be estimated jointly with resource cost. A semantically useful token may have lower net value if retaining it causes substantial KV-cache pressure, memory bandwidth contention, or batching inefficiency. Conversely, a moderately useful token may have high net value if it enables cache reuse or avoids additional prefill computation. Future value functions should therefore incorporate memory footprint, prefill--decode asymmetry, and accelerator utilization, rather than treating semantic contribution alone as value~\cite{kwon2023vllm,agrawal2024sarathi,zhong2024distserve,liu2024kivi}.

\textbf{Online Bayesian updating.}
Rather than assigning a fixed score to each token, a serving system can maintain a distribution over token value. As decoding proceeds, evidence from attention patterns, retrieval use, verifier feedback, partial reasoning progress, or cache behavior can update this distribution. This would allow the system to distinguish tokens that are confidently low-value from tokens whose value is uncertain but potentially high.

\textbf{Decision-specific value models.}
Different serving decisions require different notions of value. Cache eviction depends heavily on future reuse and memory pressure; pricing depends on utility, transparency, and incentives; routing depends on expected quality-cost trade-offs; and reasoning-budget allocation depends on the marginal benefit of additional test-time computation. A promising direction is therefore to maintain multiple decision-specific value models rather than a single universal token-value estimator.

\textbf{Transparent and auditable accounting.}
For user-facing services, token accounting should distinguish among input tokens, output tokens, hidden reasoning tokens, retrieved-context tokens, and tool-use tokens. This does not require revealing proprietary model internals, but it does require clearer accounting of what kinds of tokenized computation were used. Verifiable or auditable inference mechanisms may become important when token accounting affects billing, compliance, or service guarantees~\cite{sun2025coin,ong2025toploc}.

\textbf{Integration with reasoning-time allocation.}
Reasoning-oriented models and test-time scaling methods suggest that output quality can depend strongly on how much computation is allocated to intermediate reasoning, search, or verification~\cite{wang2023selfconsistency,yao2023tree,snell2024scaling,deepseek2025r1,team2025kimi}. Real-time token value accounting should therefore extend beyond visible input and output tokens to include reasoning tokens and verification tokens. The central question is not merely how many tokens were used, but whether the marginal reasoning token was worth its cost.

\section{Challenge II: Optimal Resource Allocation Under Probabilistic and Latency Constraints (Decision-Making)}

\subsection{The Economic Problem}

The second challenge concerns decision-making: once token values are estimated, possibly only approximately, how should limited computational resources be allocated across tokens, sequences, and requests? In computational token economics, this is not merely a static budgeting problem. Allocation decisions must be made online, under uncertainty, and within the latency constraints of real inference systems.

\textbf{Key question.} Given limited memory, compute, bandwidth, and time, how should an inference system allocate resources to tokens so as to maximize expected utility while satisfying real-time serving constraints?

Concrete instances of this problem include:
\begin{itemize}[leftmargin=*]
    \item \textbf{KV-cache allocation}: deciding which tokens' key--value states should be retained, compressed, quantized, offloaded, or evicted.
    \item \textbf{Reasoning-compute allocation}: deciding how many intermediate reasoning, search, or verification steps should be allocated to different requests or subproblems.
    \item \textbf{Bandwidth allocation}: deciding which activations, KV blocks, or intermediate states should be communicated across devices in distributed inference, and which should instead be recomputed or approximated.
    \item \textbf{Admission and scheduling}: deciding which requests or token streams should be admitted, delayed, batched, preempted, or routed to different models or hardware tiers.
\end{itemize}

The economic objective is to maximize expected service utility, such as answer quality, task success, revenue, or user satisfaction, subject to hard constraints on memory, latency, and accelerator capacity. Unlike classical token-pricing problems, the allocation mechanism itself must be computationally cheap enough to run inside the serving loop.

\subsection{The Computational Challenge}

A useful abstraction is to view token-level resource allocation as an online stochastic control problem. At decoding step $t$, the system observes a history
$\mathcal{H}_t = (x_{1:t}, a_{1:t-1}, r_{1:t-1}, s_t),$
where $x_{1:t}$ denotes the observed input and generated tokens up to step $t$, $a_{1:t-1}$ denotes previous allocation decisions, $r_{1:t-1}$ denotes observed rewards or proxy signals, and $s_t$ denotes system state such as available memory, batch composition, queue length, and hardware utilization. The system then chooses an allocation action
$a_t \in \mathcal{A}(s_t),$
for example retaining, compressing, evicting, recomputing, routing, or assigning additional reasoning budget.

A stylized constrained formulation is:
\begin{equation}
\begin{aligned}
\max_{\pi} \quad
& \mathbb{E}_{\pi}\!\left[ U(Y) - \sum_{t=1}^{T} C(a_t, s_t) \right] \\
\text{s.t.} \quad
& \sum_{j \in \mathcal{K}_t} m_j(a_t) \leq M_t, \qquad \forall t, \\
& \ell_t(a_t, s_t) \leq L_t, \qquad \forall t, \\
& a_t = \pi_t(\mathcal{H}_t), \qquad \forall t, \\
& \Pr_{\pi}\!\left( \ell_t(a_t, s_t) > L_t \right) \leq \delta_t, \qquad \forall t .
\end{aligned}
\label{eq:online_allocation}
\end{equation}

Here, $Y$ is the final model output, $U(Y)$ is its utility, $C(a_t,s_t)$ is the computational or monetary cost of the allocation action, $\mathcal{K}_t$ is the set of active cached tokens or token blocks at time $t$, $m_j(a_t)$ is the memory footprint of token or block $j$ under action $a_t$, $M_t$ is the available memory budget, $\ell_t$ is the decision and execution latency, and $L_t$ is the latency budget. The policy $\pi_t$ must be adapted to the observed history $\mathcal{H}_t$; it cannot depend on future tokens or future requests that have not yet arrived.

The last constraint is optional but important in production systems: latency is often not only an expectation constraint but also a tail-risk constraint. A policy with good average utility may still be unacceptable if it frequently violates per-token or per-request latency targets.

\textbf{Why the problem is hard.}

\textit{1. Future token values are uncertain.}
Many allocation decisions must be made before the system knows whether a token will become important later. A system prompt, a retrieved fact, or an early variable definition may have low immediate attention but high future value. Conversely, many recent tokens may be locally salient but economically redundant.

\textit{2. The action space is combinatorial.}
If $N$ active tokens or token blocks each have $K$ possible resource levels, the naive action space has size $K^N$. For example, choices such as full precision, compressed storage, offloading, recomputation, or eviction rapidly become intractable when applied at token granularity.

\textit{3. Decisions must be made under strict latency budgets.}
Unlike offline optimization, serving-time allocation cannot rely on expensive global solvers, long Monte Carlo rollouts, or complex market-clearing mechanisms. The allocation logic must often run at the granularity of decoding steps, batches, or scheduler iterations, where even small overheads can reduce throughput or increase tail latency.

\textit{4. Allocation and generation are coupled.}
In autoregressive generation, allocation decisions affect the model state, which affects future tokens, which in turn changes future allocation needs. This feedback loop makes the problem closer to online control than to static resource assignment.

\textit{5. Local optimality may conflict with global optimality.}
A locally rational decision, such as evicting a currently unused token, may reduce downstream quality. Conversely, preserving too many potentially useful tokens may reduce batch size, increase memory pressure, and harm system-wide throughput. This is a concrete manifestation of the broader impossibility triangle between fine-grained valuation, real-time performance, and allocation optimality.

\subsection{The Speculative Decoding Complication}

Speculative decoding adds another layer of economic complexity. In a typical speculative decoding setup, a cheaper draft model proposes candidate tokens, and a stronger target model verifies them. Accepted draft tokens reduce decoding cost, while rejected tokens consume computation without directly appearing in the final output.

From a token-economic perspective, speculative tokens are uncertain assets. Their value depends on whether they are accepted by the verifier and whether they accelerate generation without degrading output quality. Let $z$ denote a proposed draft token or draft block, $p_{\mathrm{acc}}(z \mid \mathcal{H}_t)$ its probability of acceptance, $v(z)$ the utility or cost saving if accepted, and $c_{\mathrm{draft}}(z)$ and $c_{\mathrm{verify}}(z)$ the draft and verification costs. A simplified expected net value is:
\begin{equation}
\mathrm{ENV}(z \mid \mathcal{H}_t)
=
p_{\mathrm{acc}}(z \mid \mathcal{H}_t) \, v(z)
-
c_{\mathrm{draft}}(z)
-
c_{\mathrm{verify}}(z)
+
I(z \mid \mathcal{H}_t),
\label{eq:speculative_env}
\end{equation}
where $I(z \mid \mathcal{H}_t)$ represents any additional informational value of the draft proposal, such as helping the system estimate uncertainty or choose a better routing or verification strategy.

This formulation highlights that speculative decoding is not only a systems optimization technique but also an allocation problem under uncertainty. A speculative token may be economically worthwhile even if it is not guaranteed to survive, provided that its expected acceleration or informational benefit exceeds its generation and verification cost. Conversely, aggressive speculation can waste compute when acceptance probability is low or verification becomes a bottleneck.

This perspective is consistent with the broader observation that different layers of the AI stack may assign different economic meanings to the same token. As Zhu argues, routers, agents, serving systems, and training pipelines may all act as marginal token allocators, deciding where the next unit of tokenized computation should be spent~\cite{zhu2026agentic}. Recent work on advantage-aware speculative reasoning similarly suggests that routing or speculation policies can improve the efficiency--accuracy trade-off by predicting when additional computation is likely to be valuable~\cite{maheswaran2025arbitrage}.

\subsection{Existing Approaches and Limitations}

Several families of practical policies can be interpreted as approximate solutions to the constrained allocation problem in Eq.~\eqref{eq:online_allocation}.

\textbf{Greedy value-based policies.}
These policies allocate resources to tokens or requests with the highest current value estimate. They are simple and fast, but they are myopic: they typically ignore future value, opportunity cost, and uncertainty in downstream generation.

\textbf{Recency-based policies.}
A common heuristic is to prioritize recent tokens and evict older ones. This is computationally attractive and often works well for local coherence. However, it is economically crude. Important instructions, retrieved evidence, or earlier definitions may remain valuable long after they first appear.

\textbf{Attention-based policies.}
Another approach is to use attention scores as a proxy for token importance. Such policies are more model-aware than pure recency heuristics, but attention is an imperfect economic signal. It is often backward-looking, layer-dependent, and not necessarily aligned with downstream utility or future marginal value.

\textbf{Predictive-importance policies.}
Learned predictors can estimate which tokens are likely to matter for future computation or output quality. For example, TokenButler studies predictable token importance for KV-cache management~\cite{abdelfattah2025tokenbutler}. Such approaches are promising because they move from reactive heuristics toward amortized value estimation. Their limitations are that they require training data, may be sensitive to distribution shift, and may optimize proxy objectives that are not identical to economic utility.

\textbf{Scheduler-level policies.}
Batching, routing, prefill--decode scheduling, and admission control policies allocate resources at the request or sequence level rather than the individual-token level. These policies are compatible with real serving systems, but they may be too coarse to exploit fine-grained token heterogeneity.

Overall, existing approaches tend to occupy different points of the impossibility triangle. Fine-grained predictive methods can improve allocation quality but add overhead. Simple heuristics satisfy latency constraints but may ignore economically important information. Scheduler-level approaches are deployable but often lack token-level optimality.

\subsection{Open Problems}

\begin{enumerate}[leftmargin=*]
    \item \textbf{Fast approximate optimization.}
    Can we design online allocation algorithms whose overhead is sublinear or near-linear in the number of active tokens or token blocks, while still providing useful approximation or regret guarantees under realistic serving constraints?

    \item \textbf{Uncertainty-aware token valuation.}
    How can systems estimate distributions over future token value, rather than only point estimates? Useful signals may include model confidence, entropy, retrieval provenance, task type, historical attention patterns, or lightweight learned predictors.

    \item \textbf{Tail-latency-constrained allocation.}
    How should allocation policies be optimized when latency constraints are hard or probabilistic rather than merely average-case? This is especially important for interactive applications, where tail latency can dominate perceived quality.

    \item \textbf{Joint allocation across memory, compute, and bandwidth.}
    Current heuristics often optimize one resource in isolation, such as KV-cache memory. Real inference systems face coupled constraints: compressing a cache entry may save memory but increase compute; offloading may save GPU memory but consume bandwidth; recomputation may reduce storage but increase latency.

    \item \textbf{Policy learning under distribution shift.}
    Can learned allocation policies generalize across models, context lengths, domains, and hardware configurations? How can such policies be audited, debugged, and safely updated in production?

    \item \textbf{Mechanism design for multi-tenant inference.}
    In shared serving environments, token allocation also raises fairness and pricing questions. How should scarce decode bandwidth, KV-cache capacity, or reasoning budget be allocated across users, agents, or applications with different values and service-level objectives?

    \item \textbf{Verifiable allocation and accounting.}
    If users are charged for visible tokens, hidden reasoning tokens, or speculative computation, how can the system provide transparent and verifiable accounting without exposing proprietary model internals or adding excessive overhead?
\end{enumerate}

\subsection{Promising Directions}

\textbf{Threshold-based policies with dynamic adjustment.}
A practical direction is to allocate expensive resources only to tokens whose estimated marginal value exceeds a dynamic threshold. The threshold can adapt to current memory pressure, queue length, batch composition, or latency slack. Such policies are simple, often $O(N)$ over active tokens or blocks, and naturally compatible with online serving.

\textbf{Hierarchical allocation.}
Instead of making fully token-level decisions everywhere, systems can first allocate resources at coarse granularity, such as request, segment, document, or KV block level, and then refine decisions only within high-priority regions. This reduces the decision space while preserving fine-grained control where it is most valuable.

\textbf{Limited lookahead.}
Purely myopic policies ignore near-term dependencies, while full-horizon planning is infeasible. A middle ground is limited lookahead over a small number of decoding steps, scheduler iterations, or speculative draft blocks. This can capture short-range dependencies without requiring global optimization.

\textbf{Amortized predictive models.}
Lightweight predictors can estimate token or block importance using features already available in the serving stack, such as attention statistics, position, retrieval metadata, cache reuse, or uncertainty signals. The cost of prediction can be amortized across batches, layers, or repeated request patterns.

\textbf{Primal--dual and budget-tracking methods.}
Online constrained optimization suggests maintaining shadow prices for scarce resources such as GPU memory, decode bandwidth, or verification capacity. When a resource becomes scarce, its shadow price rises, making the policy more selective. This provides an economic interpretation of adaptive resource management while remaining computationally lightweight.

\textbf{Hardware-aware allocation primitives.}
Because allocation decisions are made inside latency-critical paths, they should be co-designed with system architecture. Useful primitives may include efficient priority queues for KV blocks, low-overhead importance counters, cache layouts that support fast eviction or compression, and scheduler interfaces that expose resource prices to higher-level policies.

\textbf{Speculation-aware routing.}
For speculative decoding and multi-model serving, routers can estimate the expected net value of drafting, verification, or escalation to a stronger model. The goal is not simply to maximize acceptance rate, but to maximize utility per unit of latency, memory, and compute.

\subsection{When Fine-Grained Token Economics Fails}
\label{subsec:when_fails}

The preceding sections have argued for finer-grained token valuation and allocation. However, a critical and often neglected question is: \emph{when should a system deliberately avoid token-level economic optimization?} Recognizing the boundaries of the approach is essential for practical deployment and for avoiding the trap of ``over-engineering'' a mechanism whose overhead exceeds its benefit. We identify three regimes in which coarse-grained or heuristic allocation is preferable to fine-grained token economics.

\paragraph{Low-heterogeneity workloads.}
When the input distribution consists predominantly of short, homogeneous prompts---for example, simple factual queries or single-turn classification tasks---the marginal value variance across tokens is small. In such regimes, the potential gain from discriminating among individual prompt tokens or output tokens is bounded by the low ceiling of task complexity. The sensing, estimation, and optimization overhead required for token-level valuation (even amortized) can consume a larger fraction of the total inference budget than the savings it produces. Empirically, if the coefficient of variation of per-token marginal values satisfies $\sigma_v / \mu_v < \epsilon_{\text{sys}}$ for a system-dependent threshold, request-level or sequence-level budgeting is Pareto-superior.

\paragraph{High batch-pressure regimes.}
Modern LLM serving achieves throughput via aggressive kernel fusion, regular memory access patterns, and uniform batching. Token-level differentiation---for example, assigning different cache eviction priorities, precision levels, or routing decisions to individual tokens within a batch---can destroy the very regularity on which kernel efficiency depends. When batch size $B$ is large and the serving bottleneck is memory-bandwidth saturation or kernel occupancy, the hardware-level efficiency loss from irregular per-token treatment can dominate the algorithmic gain from better allocation. In this regime, batch-homogeneous policies (uniform compression, uniform retention) may maximize goodput even if they retain some low-value tokens.

\paragraph{Hard latency floors.}
In applications with strict tail-latency requirements---such as real-time speech transcription, code autocomplete, or interactive gaming---any additional per-token decision logic is infeasible regardless of its expected accuracy. If the 99th-percentile latency budget $L_{99}$ is smaller than the combined cost of value estimation and allocation optimization ($T_{\text{sense}} + T_{\text{alloc}} > L_{99} - L_{\text{gen}}$, where $L_{\text{gen}}$ is the bare generation latency), then token economics must be reduced to pre-computed, coarse heuristics or abandoned entirely. This is the most unforgiving regime: the trilemma collapses to a single axis (real-time execution), and both granularity and optimality are sacrificed by design.

\begin{figure}[t]
\centering
\includegraphics[width=0.52\textwidth]{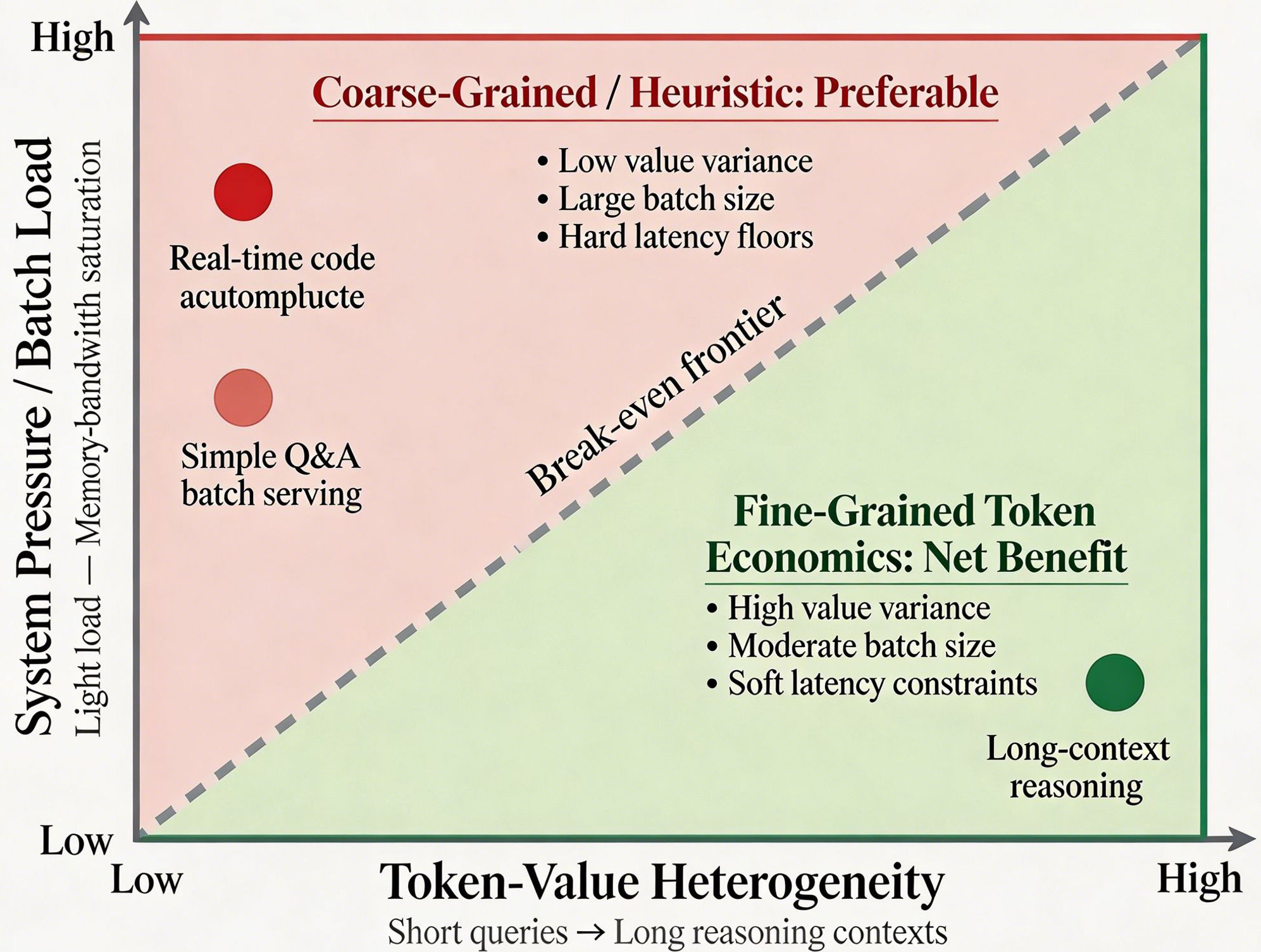}
\caption{Regimes where fine-grained token economics is advantageous (green) versus regimes where coarse-grained or heuristic allocation is preferable (red). The x-axis captures token-value heterogeneity (e.g., long-context reasoning vs.\ short queries); the y-axis captures system pressure (e.g., batch size, memory bandwidth saturation, latency constraints). The diagonal frontier indicates that even high-heterogeneity workloads may not justify token-level optimization under extreme serving pressure.}
\label{fig:fine_grained_boundary}
\end{figure}

These three regimes are not pathological edge cases; they describe large fractions of production traffic. A practical computational token economics framework must therefore include a \emph{meta-decision layer} that selects the appropriate economic granularity based on workload characteristics, system load, and latency constraints. Figure~\ref{fig:fine_grained_boundary} illustrates this trade-off space.

The existence of these failure regimes has methodological implications. Benchmarks that evaluate token-economic mechanisms should report not only average-case utility gains, but also the \emph{break-even point}---the workload and system conditions under which the mechanism's overhead begins to dominate its benefit. Without such negative-result characterization, the field risks developing solutions in search of problems.

\section{Challenge III: System Architecture for Economic Efficiency}

\subsection{The Economic Problem}

The first two challenges focus on how token value can be estimated and how scarce resources can be allocated under computational constraints. A third challenge is architectural: even if value estimates and allocation rules are available, current LLM serving systems are not designed to expose the right control points for implementing them.

\textbf{Key question.} How should LLM inference systems be architected so that token-economic mechanisms can be implemented with low overhead, predictable latency, and auditable behavior?

\textbf{Architectural objective.} The goal is not to replace inference engines with fully general economic markets. Rather, the goal is to make economic signals first-class system inputs. An economic-aware inference system should allow token values, request priorities, latency budgets, memory pressure, and verification requirements to influence scheduling, batching, routing, KV-cache management, and approximation decisions.

This perspective turns the serving stack from a purely throughput-oriented token processing pipeline into a resource allocation system in which compute, memory, bandwidth, and verification effort are assigned according to economically meaningful signals. The main difficulty is that these signals must be acted upon within the tight timing constraints of online inference.

\subsection{The Computational Challenge}

Modern LLM serving systems are primarily optimized for throughput, latency, and hardware utilization. Techniques such as continuous batching, prefill--decode scheduling, KV-cache management, tensor parallelism, pipeline parallelism, and speculative decoding are designed to improve system efficiency under workload constraints. However, these mechanisms usually treat tokens and requests according to system-level criteria, such as sequence length, arrival time, batch compatibility, and memory footprint, rather than explicit economic value.

This creates an architectural mismatch. Token-economic mechanisms often require differentiated treatment of tokens or requests, while production inference systems rely on regularity, batching, and amortization. A mechanism that is appealing at the economic level may therefore become infeasible if it requires frequent reconfiguration, per-token counterfactual evaluation, global coordination, or fine-grained hardware control during decoding.

An economic-aware architecture must therefore satisfy four requirements.

\begin{enumerate}[leftmargin=*]
    \item \textbf{Value-aware control points.}
    The serving stack must expose places where economic information can affect decisions, such as admission control, routing, batching, prefill scheduling, decode scheduling, KV-cache retention, approximation level, and verification policy.

    \item \textbf{Low-overhead signaling.}
    Economic signals must be represented in a compact form and propagated through the runtime without dominating inference cost. In practice, this suggests using coarse value classes, priorities, budgets, or shadow prices rather than full per-token optimization at every layer.

    \item \textbf{Hardware-aware implementation.}
    Economic policies must respect the discrete and coupled nature of accelerator resources. Compute units, memory bandwidth, cache capacity, and interconnect bandwidth cannot be arbitrarily subdivided at token granularity without overhead.

    \item \textbf{Auditable and verifiable accounting.}
    If tokens are used as units of pricing, billing, or resource accounting, the system should provide sufficient evidence that token counts, allocation policies, and charging rules were applied as claimed.
\end{enumerate}

These requirements reinforce the impossibility triangle introduced earlier. Finer-grained economic control can improve allocation quality, but it increases signaling, scheduling, and verification overhead. Real-time performance pushes the system toward local and approximate decisions. Stronger optimality requires more complete information and coordination, which may be unavailable at serving time.

\subsection{Specific Architectural Challenges}

\subsubsection{Hardware-Aware Economic Control}

\textbf{Problem.} Accelerator resources are discrete, shared, and highly coupled. GPU streaming multiprocessors, tensor cores, memory bandwidth, on-chip memory, KV-cache storage, and interconnect links cannot be assigned to tokens in the same way that an abstract economic model assigns divisible resources.

Let $v_i$ denote an estimated value signal for token or request $i$, and let $r_i$ denote the vector of resources assigned to it. A simplified economic objective may be written as
\begin{equation}
    \max_{\{r_i\}} \sum_i v_i \, q_i(r_i)
    \quad
    \text{s.t.}
    \quad
    \sum_i r_i \leq R,
\end{equation}
where $q_i(r_i)$ represents the quality, utility, or service contribution obtained from allocating resources $r_i$, and $R$ denotes the available system capacity. In real inference systems, however, this formulation is only an abstraction. The feasible set is shaped by batching constraints, kernel launch overheads, memory layout, parallelism strategy, and interference among concurrent requests.

\textbf{Complications.}
First, accelerator resources are not continuously divisible. A scheduler can prioritize requests, alter batch composition, choose approximation modes, or evict cache entries, but it cannot generally allocate an arbitrary fractional amount of an SM or memory channel to an individual token. Second, tokens and requests interfere through shared memory bandwidth, cache occupancy, and synchronization. Third, changing an allocation policy may itself introduce overhead, reducing the benefit of economic optimization.

The architectural question is therefore not simply whether high-value tokens should receive more resources. Rather, the question is which control knobs provide sufficient economic differentiation without destroying hardware efficiency. Examples include priority-aware batching, value-aware KV-cache retention, adaptive precision, selective verification, and routing to models or devices with different cost--quality profiles.

\subsubsection{Economic-Aware KV-Cache and Memory Management}

KV-cache memory is one of the clearest places where token economics meets system architecture. Input tokens consume context-window capacity and KV-cache storage, while long contexts can impose significant memory pressure during decoding. However, not all context tokens contribute equally to downstream utility. Some tokens may be essential instructions, constraints, or retrieved evidence, while others may be redundant or weakly relevant.

An economic-aware memory manager would treat KV-cache capacity as a scarce resource. Given token-level or block-level importance estimates, the system could decide which cache entries to retain, compress, offload, recompute, or evict. This connects naturally with recent work on KV-cache compression and selection, which already shows that inference systems can exploit non-uniform token importance from a systems perspective.

The challenge is that memory decisions are path-dependent. Evicting or compressing a token may reduce future quality, but the future usefulness of that token is uncertain at the time of the decision. Moreover, cache policies must be implemented with predictable latency and without excessive metadata overhead. This makes KV-cache management a central test case for computational token economics: it requires valuation, online allocation, and hardware-aware implementation simultaneously.

\paragraph{Example: Value-Aware PagedAttention.}
To make the abstract challenge concrete, consider extending vLLM's PagedAttention~\cite{kwon2023vllm} with block-level economic signals. In standard PagedAttention, the KV cache is partitioned into fixed-size pages (blocks). Eviction and retention decisions are typically driven by system-level heuristics such as recency, reference counting, or uniform compression. A value-aware extension attaches a \emph{shadow price} $\lambda_b$ to each block $b$, computed from the cumulative estimated marginal value of the tokens it contains:
\begin{equation}
    \lambda_b
    =
    \frac{\sum_{i \in \mathrm{tokens}(b)} \hat{v}_i}{\mathrm{size}(b)}
    -
    \mu \cdot \mathrm{pressure}_b,
    \label{eq:block_shadow_price}
\end{equation}
where $\hat{v}_i$ is the estimated net value of token $i$ (Eq.~\eqref{eq:token_marginal_value}), $\mathrm{size}(b)$ is the memory footprint of the block, and $\mathrm{pressure}_b$ captures the contention cost of retaining $b$ (e.g., preventing batch-size expansion or causing offloading). The scheduler then evicts blocks in ascending order of $\lambda_b$, subject to the constraint that metadata maintenance and $\lambda_b$ updates add less than $\gamma$ (e.g., $1\%$) to the prefill or decode latency.

\begin{figure}[t]
\centering
\includegraphics[width=0.86\textwidth]{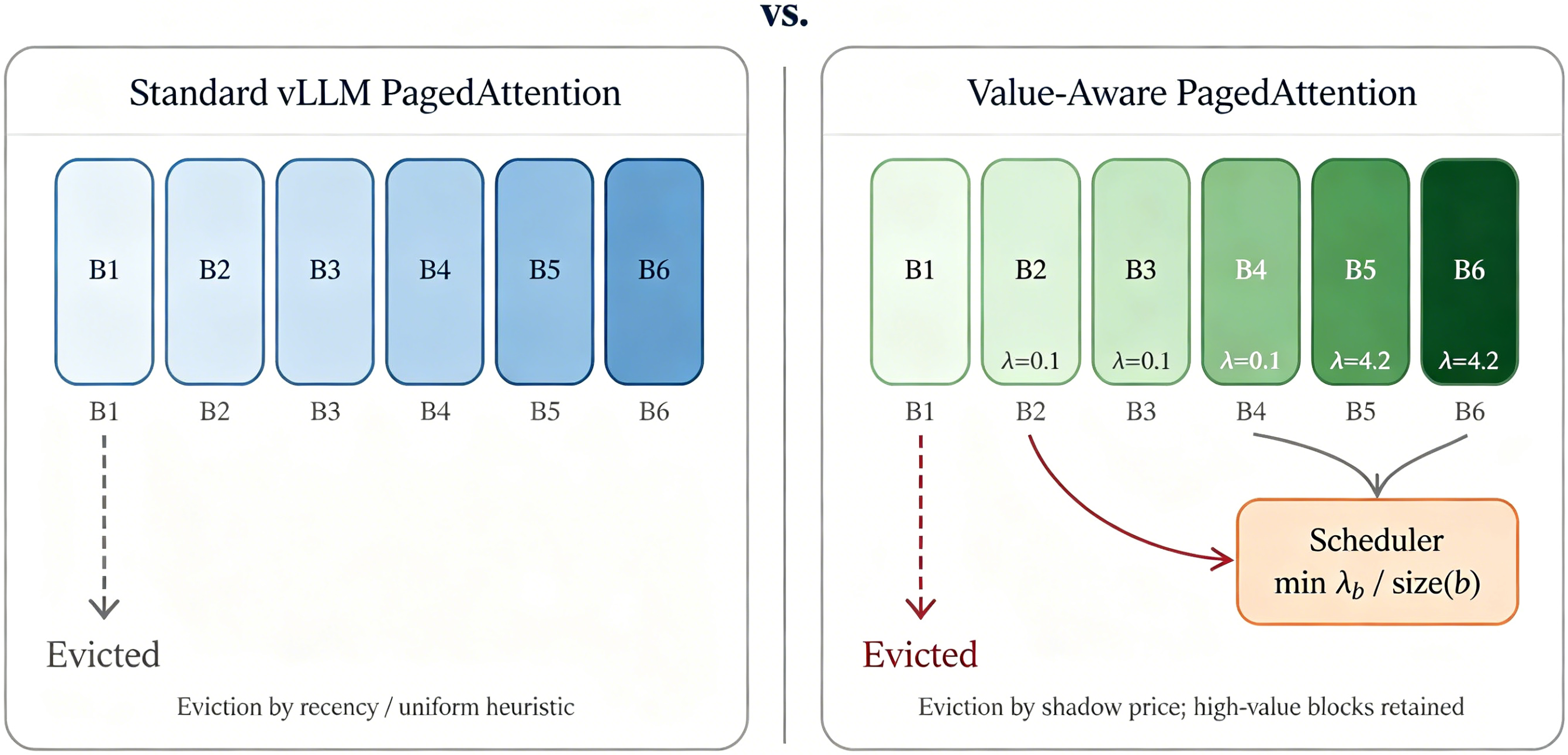}
\caption{Contrast between standard PagedAttention (left) and a Value-Aware extension (right). In the standard design, eviction is driven by recency or uniform heuristics. In the value-aware design, each KV block carries a shadow price $\lambda_b$; the scheduler evicts the block with the lowest value-to-cost ratio, retaining high-value context (e.g., system prompts, retrieved evidence) while releasing low-value padding or redundant history.}
\label{fig:value_aware_pagedattention}
\end{figure}

This design illustrates the trilemma in microcosm. \textbf{Granularity} is relaxed from token-level to block-level, reducing the effective decision space to the number of pages rather than individual tokens and keeping metadata overhead bounded. \textbf{Real-time execution} is preserved because block-level scoring can be amortized across scheduler iterations and updated incrementally when tokens are generated or evicted. \textbf{Optimality} is approximate: the policy is greedy over blocks rather than solving a global knapsack over tokens, but the shadow-price formulation provides an economic interpretation of the eviction order. Figure~\ref{fig:value_aware_pagedattention} contrasts the standard and value-aware architectures.

\subsubsection{Distributed Economic Coordination}

\textbf{Problem.} Large-scale inference often spans multiple devices or nodes through tensor parallelism, pipeline parallelism, expert parallelism, or disaggregated prefill and decoding. Economic decisions made at one component may affect costs and bottlenecks elsewhere.

\textbf{Complications.}
First, each node observes only part of the serving state. A prefill worker may see context length and memory pressure, while a decode worker may experience a different bottleneck. Second, communicating detailed value information across nodes adds latency and bandwidth overhead. Third, resource prices may differ across components: memory may be scarce on one device, compute on another, and network bandwidth on a third.

This suggests that distributed token economics should avoid fully centralized optimization in the online path. A more realistic architecture may use hierarchical control: global policies determine coarse priorities, budgets, or service classes, while local schedulers make fast hardware-aware decisions using local congestion signals. Such a design sacrifices global optimality but can preserve real-time performance.

\subsubsection{Multi-Tenant Economic Isolation}

\textbf{Problem.} Shared inference services must allocate resources across users, applications, or agents with different budgets, priorities, and latency requirements. If the system accepts user-provided value signals, then strategic behavior becomes possible.

\textbf{Complications.}
Users may overstate the importance of their requests, split workloads to exploit scheduling rules, or select prompts that induce disproportionate hidden computation. Conversely, a purely provider-controlled allocation policy may be opaque to users and difficult to audit. There is therefore a tension between economic efficiency, fairness, incentive compatibility, and transparency.

A practical architecture should distinguish between internal value estimates and externally declared willingness to pay. Internal estimates can guide system efficiency, such as cache retention or approximation level. External economic signals, such as priority tiers or budgets, should be mediated by mechanisms that prevent simple manipulation. Recent work on mechanism design and pricing for generative AI services is relevant here; for example, token caps, service tiers, and budget-constrained allocation can be interpreted as coarse economic instruments rather than exact per-token markets~\cite{zhong2025token}.

\subsubsection{The Verifiable Computation and Accounting Problem}

\textbf{Motivation.}
If tokens are economic primitives, then token accounting becomes economically consequential. Users may be charged based on input tokens, output tokens, hidden reasoning tokens, tool-use tokens, or other internal computation. In opaque API settings, users may not be able to directly verify whether the reported token counts or internal computation traces are accurate. This creates a trust problem for token-based pricing and accounting.

Recent work has begun to study this issue. Artola Velasco et al.\ argue that pay-per-token pricing can create incentives for misreporting token usage when users cannot independently verify token counts~\cite{artola2025overcharging}. Sun et al.\ study verification problems associated with invisible or hidden reasoning tokens in commercial LLM APIs~\cite{sun2025coin}. More broadly, trustless or verifiable inference methods suggest that token-level economic mechanisms may need to be coupled with proof systems or audit mechanisms.

\textbf{Economic architecture requirement.}
An economic-aware inference system should provide evidence for at least three kinds of claims:
\begin{enumerate}[leftmargin=*]
    \item the number and type of tokens charged to the user;
    \item the pricing or budget rule applied to those tokens;
    \item the compliance of the serving process with declared allocation or verification policies.
\end{enumerate}

\textbf{Computational challenge.}
Verification is not free. Let $C_{\mathrm{inf}}$ denote the baseline inference cost and $C_{\mathrm{ver}}$ denote the additional cost of producing and checking evidence. A verification mechanism is practical only when
\begin{equation}
    \frac{C_{\mathrm{ver}}}{C_{\mathrm{inf}}}
\end{equation}
is small enough for the target service-level objective. The acceptable overhead depends on the application: high-stakes or high-value inference may tolerate larger verification costs, while low-latency consumer services may not.

\textbf{Potential approaches.}
Several architectural directions are possible:

\begin{itemize}[leftmargin=*]
    \item \textbf{Commitment schemes for token accounting.}
    Merkle-tree or hash-chain commitments can bind the provider to a sequence of tokens, token blocks, or accounting events. Such methods can support later auditing, but they do not by themselves prove semantic correctness or faithful model execution.

    \item \textbf{Proofs or fingerprints of inference execution.}
    Lightweight proof or fingerprinting methods can provide evidence that an inference trace is consistent with a claimed computation. For example, locality-sensitive hashing or embedding-based fingerprints may reduce storage relative to storing full activations or embeddings~\cite{ong2025toploc}. The trade-off is that such methods may provide probabilistic rather than full cryptographic guarantees, depending on their construction.

    \item \textbf{Trusted execution environments.}
    TEEs can protect parts of the inference process and produce attestations about executed code. They may be useful for accounting and policy compliance, but deployment complexity, hardware availability, performance overhead, and side-channel risks remain important concerns.

    \item \textbf{Incentive-compatible pricing rules.}
    Another approach is to reduce the incentive to misreport token counts. For instance, alternative pricing units or coarse service tiers may weaken the dependence of revenue on unverifiable internal token accounting~\cite{artola2025overcharging}.
\end{itemize}

These approaches differ in security assumptions, overhead, auditability, and compatibility with existing serving stacks. The broader lesson is that verifiability must be treated as part of the system architecture, not as an external billing feature.

\subsection{Open Problems}

\begin{enumerate}[leftmargin=*]
    \item \textbf{Value-aware scheduling abstractions.}
    What is the right interface between economic policies and inference schedulers? Should policies expose scalar priorities, value classes, latency budgets, shadow prices, or richer utility functions?

    \item \textbf{Economic-aware GPU kernels and runtimes.}
    Can attention, KV-cache, routing, and decoding kernels incorporate economic signals without harming occupancy, memory coalescing, or batching efficiency?

    \item \textbf{Online pricing of scarce system resources.}
    Can serving systems maintain real-time estimates of the marginal cost of memory, bandwidth, decoding slots, and verification effort, and use these estimates to guide allocation?

    \item \textbf{Distributed economic control.}
    How should value and congestion signals be propagated across disaggregated prefill workers, decode workers, routers, and cache servers without introducing excessive synchronization overhead?

    \item \textbf{Verifiable token accounting.}
    What level of evidence is sufficient for different applications: exact cryptographic proofs, probabilistic audits, trusted hardware attestations, or transparent accounting logs?

    \item \textbf{Fairness and incentive compatibility.}
    How can shared inference systems prevent strategic manipulation while still allowing users or applications to express heterogeneous values, priorities, and budgets?

    \item \textbf{Hardware support for token economics.}
    Should future AI accelerators expose primitives for priority-aware scheduling, cache partitioning, accounting counters, secure metering, or low-overhead attestation?
\end{enumerate}

\subsection{Promising Directions}

\textbf{Layered economic architecture.}
A promising direction is to separate economic decision-making across layers. The application layer specifies high-level objectives, such as budget, latency, quality, or verification requirements. The runtime layer translates these objectives into scheduling, routing, batching, and cache-management decisions. The kernel layer implements these decisions with minimal overhead. The hardware layer exposes counters, isolation mechanisms, and scheduling primitives. This separation allows economic policies to evolve without requiring every policy to be hard-coded into kernels or hardware.

\textbf{Coarse-to-fine economic control.}
Because fine-grained token-level optimization is expensive, practical systems may use economic mechanisms at coarse granularity and heuristics at fine granularity. For example, request-level budgets or service classes can determine routing and admission control, while block-level token importance can guide KV-cache compression or eviction. This design reflects the impossibility triangle: coarse decisions preserve real-time performance, while selective fine-grained control improves allocation quality where the benefit is largest.

\textbf{Economic-aware memory hierarchy.}
KV-cache capacity, memory bandwidth, and offload bandwidth should be treated as economically scarce resources. A memory hierarchy that supports value-aware retention, compression, recomputation, and eviction could provide a concrete implementation path for computational token economics. The Value-Aware PagedAttention example above illustrates one such path: block-level shadow prices (Eq.~\eqref{eq:block_shadow_price}) allow the system to retain high-value context (e.g., system prompts, retrieved evidence) while evicting low-value padding or redundant history, without breaking the page-aligned memory layout on which kernel efficiency depends. This direction is especially attractive because token heterogeneity is already visible in context selection and KV-cache optimization, and existing compression methods (H$_2$O, SnapKV, PyramidKV) can be reinterpreted as approximate special cases of value-aware eviction with fixed value proxies.

\textbf{Transparent accounting and audit layers.}
Economic efficiency requires trust when users are charged for tokens or hidden computation. Serving systems can include an accounting layer that records token counts, pricing events, policy decisions, and verification metadata. Depending on the application, this layer may use cryptographic commitments, probabilistic audits, TEEs, or external logs. The goal is not necessarily to prove every floating-point operation, but to make economically relevant claims auditable at acceptable cost.

\textbf{Economic virtualization.}
A virtualization layer between economic policies and heterogeneous hardware could make token-economic mechanisms portable across GPUs, accelerators, and distributed serving backends. Instead of exposing device-specific details, the system would expose abstract resources such as decode capacity, KV-cache capacity, memory bandwidth, and verification budget. The runtime would map these abstractions to hardware-specific actions. This would allow the same economic policy to run across different serving architectures while preserving hardware-aware optimization.

Overall, system architecture determines whether token-economic mechanisms remain abstract models or become deployable inference mechanisms. The key architectural challenge is to expose enough control for economic differentiation while preserving the regularity, batching efficiency, and hardware utilization on which modern LLM serving depends.

\section{Broader Implications and Connections}\label{sec:broader}

\subsection{Computational Token Economics as a Research Domain}

The preceding sections suggest that token economics in large language model
systems cannot be reduced to pricing alone. Token-level decisions are executed
inside serving systems with strict latency, memory, batching, and accelerator
constraints. We therefore propose \emph{Computational Token Economics} as a
research domain concerned with the design, analysis, and implementation of
token-economic mechanisms under the computational constraints of real AI
infrastructure.

More precisely, Computational Token Economics studies how token-level value,
cost, and allocation principles can be made operational in systems where
decisions must often be approximate, online, hardware-aware, and uncertainty
aware. Its goal is not merely to define economically meaningful objectives, but
also to understand which objectives can be implemented with acceptable latency,
memory overhead, and system complexity.

This perspective is organized around four principles.

\begin{enumerate}[leftmargin=*]
    \item \textbf{Economic grounding.}
    Token-level decisions should be connected to value, cost, utility, or
    marginal contribution, rather than relying only on task-agnostic heuristics.
    This does not imply that exact valuation is always feasible; rather, it
    requires that approximations be interpretable in economic terms.

    \item \textbf{Computational tractability.}
    Valuation and allocation mechanisms must respect the time scale of LLM
    serving. In practice, this favors online, approximate, amortized, or
    hardware-aware algorithms over exact procedures whose overhead would exceed
    the value they attempt to recover.

    \item \textbf{System integration.}
    Economic mechanisms should be designed together with serving architecture.
    Token allocation interacts with prefill--decode scheduling, batching,
    KV-cache placement, routing, speculative decoding, admission control, and
    accelerator utilization. As a result, economic optimality cannot be assessed
    independently of system constraints.

    \item \textbf{Empirical validation.}
    Proposed mechanisms should be evaluated on realistic workloads and system
    metrics, including latency, throughput, memory usage, quality, cost, and
    robustness under uncertainty. Economic improvements that rely on unrealistic
    valuation or coordination assumptions may not survive deployment.
\end{enumerate}

Computational Token Economics is related to several established areas, but it
differs from each in its basic object of study: tokens as heterogeneous,
interdependent, and computationally costly economic units.

\begin{table}[ht]
\centering
\caption{Computational Token Economics and Related Fields}
\label{tab:related}
\begin{tabular}{@{}p{3.6cm}p{5cm}p{6cm}@{}}
\toprule
\textbf{Field} & \textbf{Relationship} & \textbf{Key Difference} \\
\midrule
Computational Advertising
& Shares real-time allocation, pricing, and auction-like decision problems.
& Token values are often implicit, context-dependent, and interdependent rather
than directly observed bids. \\
\addlinespace
Algorithmic Game Theory
& Provides tools for mechanism design, incentives, and allocation under
strategic behavior.
& LLM serving imposes tight latency, memory, and hardware constraints, while
outcomes are probabilistic and difficult to verify exactly. \\
\addlinespace
Operating Systems
& Studies scheduling, caching, admission control, and resource management.
& Computational Token Economics adds value-aware objectives and token-specific
constraints such as context length, KV-cache pressure, and decoding cost. \\
\addlinespace
ML Systems
& Focuses on scalable and efficient model training and inference.
& The emphasis here is not only on performance, but also on value-aware
allocation, accounting, pricing, and economic transparency. \\
\addlinespace
Computational Economics
& Provides models for allocation, optimization, and decision-making under
constraints.
& The mechanisms must be embedded in stochastic, high-throughput AI serving
systems where exact global optimization is typically infeasible. \\
\bottomrule
\end{tabular}
\end{table}

This positioning also clarifies the role of the Token Economics Impossibility
Triangle introduced earlier. Fine-grained token valuation, real-time execution,
and strong allocation optimality are individually desirable, but jointly
difficult to achieve in online LLM serving. Computational Token Economics
therefore studies not only optimal mechanisms, but also the approximations,
interfaces, and architectural supports that make such mechanisms deployable.

Finally, we note that Computational Token Economics as defined here 
is deliberately scoped to the \emph{inference-system boundary}. 
Broader questions---market structure (monopoly platforms, token 
futures), multi-agent non-cooperative allocation, and security 
externalities (adversarial token consumption, prompt-injection 
costs)---are important complements, but they require game-theoretic 
and institutional analysis beyond the computational focus of this 
paper.

\subsection{Impact on AI Development Practices}

If tokens are treated as economic primitives, AI development practices may need
to evolve beyond conventional performance-centric optimization. Current serving
systems are commonly evaluated by throughput, latency, hardware utilization, and
quality metrics. These remain essential, but they may be insufficient when
different tokens carry different costs, values, and downstream effects.

A first implication is a shift from maximizing raw throughput to improving
\emph{value per unit cost}. For example, a system may prefer to allocate more
tokens to requests where additional reasoning, retrieval, or verification is
likely to improve utility, while reducing redundant context or low-value
generation elsewhere. Such policies require approximate value estimation, but
they also require safeguards against excessive overhead: the cost of estimating
token value should not dominate the benefit of better allocation.

A second implication is the need for new evaluation metrics. Relevant metrics
may include value-normalized latency, utility per dollar, marginal quality gain
per generated token, KV-cache value density, and robustness of allocation
decisions under uncertain token importance. In multi-user or multi-agent
settings, additional metrics may be required to evaluate fairness, strategic
robustness, and the distribution of benefits across participants.

A third implication concerns the development workflow. Economic mechanisms
should not be added only as an external pricing layer after the serving system
has been built. Instead, model design, inference algorithms, routing policies,
memory management, and accounting interfaces may need to be co-designed. This
is especially important for hidden reasoning tokens, tool-use traces, and
intermediate agent computations, where the economically relevant token
consumption may not be directly visible to users.

Finally, deployment monitoring may need to include economic observability in
addition to system observability. Operators may need to track not only latency
and errors, but also token-level cost attribution, value estimates, allocation
decisions, and discrepancies between expected and realized utility. Such
monitoring is a prerequisite for debugging, auditing, and improving
token-economic mechanisms in production systems.

\subsection{Societal and Ethical Considerations}

Economic optimization in AI systems raises broader societal and ethical
questions. If access to higher-quality reasoning, longer context, or more
verification is governed primarily by willingness or ability to pay, then
token-economic mechanisms could amplify existing inequalities in access to AI
capabilities. Conversely, ignoring cost and scarcity may make large-scale AI
services economically unsustainable. The challenge is therefore not whether
economic allocation should exist, but how it should be designed with
appropriate safeguards.

One concern is basic access. Systems may need mechanisms that preserve a
minimum level of service quality for low-budget users, educational use,
accessibility needs, or other socially valuable applications. This could take
the form of quotas, subsidies, priority rules, or service tiers. However, such
mechanisms must be specified carefully, because they interact with congestion,
latency, and resource allocation in the same infrastructure used by all users.

A second concern is transparency. Users should be able to understand, at an
appropriate level of abstraction, how token usage affects cost, latency, and
quality. This is particularly important when systems use hidden reasoning
tokens, retrieval steps, tool calls, or internal agent communication. Without
transparent accounting, users may be unable to assess whether charges are
reasonable or whether the system followed the advertised policy.

A third concern is accountability and auditability. Providers may need to show
that allocation, pricing, and accounting mechanisms are consistent with stated
rules. Regulators or third-party auditors may also require evidence that
certain classes of users are not systematically disadvantaged by economic
allocation policies. Recent work on transparent token accounting and verifiable
inference suggests possible technical foundations for such accountability,
although practical deployment remains an open problem~\cite{sun2025coin,ong2025toploc}.

These considerations indicate that Computational Token Economics should not be
understood only as a tool for efficiency. It also provides a language for
discussing fairness, transparency, and governance in AI systems where tokens are
scarce, valuable, and increasingly consequential.

\subsection{Long-term Vision}

The long-term vision is that future AI systems may treat tokens as first-class
economic and computational objects. In such systems, token allocation would be
guided by explicit estimates of value and cost; users could express
fine-grained preferences over quality, latency, privacy, and expenditure;
serving systems could allocate scarce memory and compute to high-value uses;
and multi-agent systems could coordinate through transparent accounting or
market-like mechanisms.

This vision should not be interpreted as a claim that fully optimal token
markets are immediately achievable. The impossibility triangle developed in
this paper suggests the opposite: fine-grained valuation, real-time execution,
and strong optimality cannot generally be maximized simultaneously under
realistic serving constraints. The practical path forward is therefore likely
to involve approximate valuation, boundedly rational allocation, hardware-aware
mechanism design, and verifiable accounting interfaces.

The research agenda outlined in Section~8 represents initial steps toward this
goal. Important directions include approximate token valuation, online
allocation under uncertainty, transparent accounting for visible and hidden
tokens, verifiable inference, and economic-aware serving architectures. Progress
on these problems would help bridge the gap between token-economic theory and
deployable AI infrastructure, making it possible to reason about tokens not
only as linguistic units or billing units, but as operational units of value,
cost, and computation.

\section{Research Agenda and Open Problems}

Computational token economics raises a set of research problems that cannot be
resolved by economic modeling or systems optimization alone. The central
difficulty is that token-level decisions must often be made before their true
utility is known, while inference systems simultaneously face strict constraints
on latency, memory bandwidth, KV-cache capacity, batching, and accelerator
utilization. This section outlines a research agenda organized around three
levels: near-term methodological problems, system-level integration challenges,
and longer-term theoretical foundations.

\subsection{Near-Term Priorities}

\textbf{  \hspace{0.16in}P1: Lightweight token-value proxies}. 
A first priority is to develop value proxies that are cheap enough to be used
inside online serving systems. Exact token attribution is generally too
expensive for real-time inference, especially when marginal contribution is
defined by counterfactual removal, Shapley-style attribution, or downstream
quality changes. Practical proxies may therefore rely on signals such as
attention structure, position, retrieval scores, uncertainty estimates,
compression sensitivity, or task-specific priors. A key open question is how to
evaluate these proxies: correlation with offline attribution is useful, but may
not be sufficient if the proxy is ultimately used for allocation, caching, or
routing. Future work should therefore report both statistical fidelity and
system-level utility.

\textbf{P2: Online allocation under uncertainty}. 
Token allocation in LLM serving is inherently online. The value of additional
context, reasoning tokens, tool calls, or verification steps is often unknown at
the moment when scheduling and memory decisions must be made. This motivates
allocation algorithms that combine approximate valuation with online decision
making under resource constraints. Candidate approaches include threshold
policies, primal--dual methods, contextual bandits, receding-horizon control,
and greedy algorithms with bounded lookahead. The relevant theoretical question
is not whether one can compute an offline optimum with complete information,
but what guarantees are possible when valuations are noisy, non-stationary, and
partially observed.

\textbf{P3: Economic-aware KV-cache and context management}. 
KV-cache capacity is one of the most direct places where token economics meets
LLM systems. Context tokens differ in their memory footprint, reuse potential,
and marginal contribution to future outputs. This suggests cache-management
policies that account for token value rather than relying only on recency,
position, or sequence-level heuristics. Open problems include value-aware
eviction, compression, prefix sharing, context pruning, and admission control.
Such policies must be evaluated against strong systems baselines, since a
policy that improves estimated value but reduces batching efficiency or
accelerator utilization may fail in end-to-end serving.

\textbf{P4: Transparent accounting for visible and hidden tokens}. 
Modern LLM services may consume not only user-visible input and output tokens,
but also hidden reasoning tokens, retrieval tokens, tool-use tokens, verifier
tokens, and speculative decoding tokens. A practical accounting framework should
distinguish these categories without assuming that all internal traces are
observable to the user. Important questions include how to expose meaningful
cost and quality signals, how to audit token consumption, and how to prevent
economic mechanisms from encouraging opaque or strategically inflated internal
computation.

\subsection{System-Level Research Challenges}

\textbf{\hspace{0.22in}S1: Economic-aware serving architecture}. 
Token-economic mechanisms cannot be designed independently of serving
architecture. Routing, batching, prefill--decode scheduling, KV-cache placement,
speculative decoding, and admission control all affect the feasible set of
economic decisions. A major systems challenge is to expose just enough
economic information to improve allocation while preserving the throughput and
latency benefits of existing inference stacks. This calls for interfaces
between value estimators, schedulers, memory managers, and model routers.

\textbf{S2: Distributed token economics}. 
Large-scale inference often spans multiple accelerators, model replicas, and
serving regions. Economic coordination in such settings must account for
communication cost, queueing effects, heterogeneous hardware, and partial
observability. Open problems include distributed value aggregation, local versus
global allocation policies, incentive-compatible routing among model replicas,
and economic admission control under load. The central challenge is to improve
resource allocation without introducing coordination overhead that dominates
the benefit of the economic mechanism itself.

\textbf{S3: Verifiable and auditable inference economics}. 
As token-level pricing and accounting become more important, users and service
providers may require mechanisms for verifying that reported token usage,
reasoning effort, or allocation decisions are consistent with stated policies.
Recent work on verifiable inference and trustless computation suggests that
cryptographic or protocol-level tools may become relevant, but their cost must
be reconciled with the latency constraints of LLM serving. Open questions
include what should be verified, at what granularity, and whether approximate
or sampled verification can provide useful guarantees at acceptable overhead.

\textbf{S4: Economic-aware training and adaptation}. 
Another direction is to train or adapt models so that they are easier to serve
under economic constraints. This may include value-weighted losses, objectives
that penalize unnecessary context dependence, distillation of value predictors,
or reinforcement-learning objectives that trade off quality, latency, and token
cost. Care is needed: optimizing directly for short-term token efficiency may
harm robustness, calibration, or long-horizon reasoning. A useful research
direction is therefore to study when economic regularization improves
deployment efficiency without degrading task performance or safety.

\subsection{Foundational Questions}

\textbf{\hspace{0.22in}F1: Theoretical limits of the token-economics triangle}. 
The impossibility triangle proposed in this paper suggests that fine-grained
valuation, real-time execution, and allocation optimality cannot generally be
maximized simultaneously. A formal theory should identify which combinations
are impossible, which are achievable under restricted assumptions, and how the
trade-off changes with model architecture, workload distribution, and hardware
constraints. Relevant tools may include online algorithms, approximation
algorithms, mechanism design, queueing theory, and computational complexity.

\textbf{F2: Definitions of token value}. 
There is no single universal notion of token value. A token may be valuable
because it improves prediction accuracy, reduces uncertainty, enables tool use,
preserves safety constraints, supports user utility, or improves future
reasoning. These notions need not agree. Future work should clarify the
relationship between marginal utility, attribution, price, cost, and
counterfactual contribution. In particular, value definitions should specify
the objective, the counterfactual baseline, the time horizon, and the resource
constraints under which value is measured.

\textbf{F3: Incentives and strategic behavior}. 
Once token values influence pricing, routing, caching, or service priority,
agents may have incentives to manipulate token usage. Users may craft prompts
to obtain more computation at lower apparent cost; providers may expose
accounting interfaces that are difficult to audit; multi-agent systems may
shift cost across agents or tools. This motivates incentive-compatible
mechanisms for token accounting and allocation, especially in multi-agent or
market-like inference environments.

\textbf{F4: Benchmarks and evaluation methodology}. 
The field needs benchmarks that evaluate economic efficiency rather than only
model quality or raw throughput. Such benchmarks should include diverse tasks,
long-context workloads, reasoning-intensive queries, retrieval-augmented
generation, tool use, and heterogeneous latency constraints. Metrics should
report quality, latency, memory use, accelerator utilization, token
consumption, and value-per-resource measures. Importantly, benchmarks should
distinguish offline attribution quality from online allocation performance.

\subsection{Cross-Cutting Themes}

\textbf{\hspace{0.22in}Approximation as a design principle}. 
Because exact valuation and globally optimal allocation are often infeasible,
approximation should be treated as a first-class design principle. Useful
systems will likely combine approximate value estimation, amortized computation,
hardware-aware heuristics, and selective high-fidelity evaluation for the most
important decisions.

\textbf{Hardware awareness}. 
Token economics is constrained by the physical organization of inference:
memory bandwidth, KV-cache layout, interconnects, batching behavior, and
accelerator scheduling. Future mechanisms should therefore be evaluated not
only at the algorithmic level, but also in realistic serving environments.

\textbf{Transparency and auditability}. 
Economic mechanisms can affect cost, quality, and trust. Transparent accounting
for visible tokens, hidden reasoning, retrieval, tool calls, and verification
steps is essential for interpreting prices and comparing systems. However,
transparency must be balanced against privacy, security, and proprietary model
internals. Recent work on transparent token accounting and verifiable 
inference suggests possible technical foundations~\cite{artola2025overcharging,sun2025coin,ong2025toploc}, 
although practical deployment---especially under privacy 
constraints such as differentially-private KV-cache access 
patterns---remains an open problem~\cite{luo2026shadow}.

\textbf{Open infrastructure}. 
Progress would benefit from open-source simulators, traces, reference
implementations, and evaluation protocols. Such infrastructure should allow
researchers to compare value estimators, allocation policies, cache-management
schemes, and verification mechanisms under common workloads and resource
constraints.

\subsection{Call to Action}

We encourage the community to treat token economics as a joint problem in
machine learning systems, computational economics, and AI infrastructure.
Promising directions include:

\begin{enumerate}[leftmargin=*]
    \item \textbf{Move beyond token counts}: evaluate not only how many tokens
    are used, but what marginal value they provide under resource constraints.

    \item \textbf{Design for deployability}: report latency, memory, batching,
    and accelerator-utilization effects when proposing token-level economic
    mechanisms.

    \item \textbf{Develop shared benchmarks}: create tasks and traces that
    measure economic efficiency across context selection, reasoning allocation,
    retrieval, caching, and verification.

    \item \textbf{Study negative results}: document cases where fine-grained
    valuation, sophisticated allocation, or verification mechanisms fail to
    justify their computational overhead.

    \item \textbf{Build interdisciplinary collaborations}: combine expertise in
    LLM inference systems, online algorithms, mechanism design, hardware
    architecture, and applied economics.
\end{enumerate}

The long-term goal is not merely to attach prices to tokens, but to understand
how token-level value, cost, and allocation can be computed under the
constraints of real AI infrastructure. This makes computational token economics
a natural bridge between economic theory and the design of scalable, trustworthy
LLM systems.

\section{Conclusion}

This paper has introduced \textbf{Computational Token Economics} as an emerging research area at the intersection of token economics, large language model serving, and computational resource allocation. While recent work increasingly treats tokens as economic primitives, we have argued that the practical realization of token-economic mechanisms depends critically on the computational constraints of real AI infrastructure. In production inference systems, token-level decisions are shaped not only by abstract notions of value and price, but also by latency budgets, memory bandwidth, KV-cache capacity, batching behavior, accelerator utilization, hidden reasoning tokens, verification overhead, and uncertainty.

The main thesis of this paper is that computational token economics is governed by a fundamental three-way tension among token-level granularity, real-time execution, and allocation optimality. We formalized this tension as the \textbf{Token Economics Impossibility Triangle}. Fine-grained valuation can improve allocation quality, but it introduces sensing, attribution, and optimization overhead. Real-time inference requires local, approximate, and hardware-aware decisions, but such decisions may ignore economically relevant global information. Strong optimality guarantees require accurate valuations and coordination across requests, models, and system resources, which are often unavailable under online serving constraints. This triangle explains why token-economic mechanisms cannot be designed independently of the systems on which they are deployed.

Building on this framework, we organized the central computational challenges into three categories. First, real-time token value accounting requires lightweight, transparent, and uncertainty-aware methods for estimating the marginal value and cost of visible, retrieved, generated, and hidden reasoning tokens. Second, constrained token allocation requires online decision mechanisms that balance quality, latency, cost, and fairness under stochastic workloads and limited resources. Third, economic-aware system architecture requires serving stacks, schedulers, memory managers, routers, and verification mechanisms to be co-designed with token-level economic objectives rather than optimized solely for throughput or latency.

A key implication is that making token economics practical is not simply a matter of implementing existing economic mechanisms more efficiently. It requires new cross-layer designs in which economic objectives and computational constraints are treated jointly. Approximate valuation, amortized attribution, hardware-aware scheduling, adaptive context management, verifiable inference, and transparent token accounting should be viewed as complementary components of the same design problem. The goal is therefore not merely to decide which vertex of the impossibility triangle to sacrifice, but to push the feasible frontier outward: achieving finer-grained economic control with lower latency and stronger allocation guarantees than current systems allow.

The emerging view of tokens as production factors, exchange media, units of account, and scarce computational resources~\cite{chen2026token,zhu2026agentic}, together with recent progress in token importance prediction~\cite{abdelfattah2025tokenbutler}, verifiable inference~\cite{sun2025coin,ong2025toploc}, and token-level pricing and incentive mechanisms~\cite{zhong2025token,artola2025overcharging}, suggests that Computational Token Economics is becoming an important foundation for future AI infrastructure. We hope that the framework developed in this paper can help clarify the design space, expose the crucial trade-offs, and motivate further research on economically grounded, computationally feasible, and trustworthy AI systems.

\bibliographystyle{IEEEtran}       
\bibliography{ReferenceFormat}       

\end{document}